\documentclass[11pt]{article}
\pdfoutput=1
\usepackage{smile}
\usepackage{setspace}

\usepackage[colorlinks,
            linkcolor=red,
            anchorcolor=blue,
            citecolor=blue
            ]{hyperref}
\usepackage{color, colortbl}
\usepackage[dvipsnames]{xcolor}
\usepackage{fullpage}
\usepackage[protrusion=true,expansion=true]{microtype}
\usepackage{tabularx}
\usepackage{float}
\usepackage{graphicx}
\usepackage{booktabs}
\usepackage{caption}
\usepackage[T1]{fontenc}
\usepackage{lmodern}

\usepackage[T1]{fontenc}
\usepackage{titletoc}
\usepackage{hyperref}
\usepackage{microtype}
\usepackage{etoolbox}

\makeatletter
\ifdefined\pdftexversion
  \@ifpackagelater{microtype}{2025/02/12}{}{%
    \def\StepRS@patch@showhyphens[#1]#2\DeclareRobustCommand\showhyphens[#3]#4{%
      \MT@exp@cs\patchcmd{showhyphens }%
        {\color@begingroup}%
        {\color@begingroup\pdfadjustspacing\z@}%
        {}%
        {\PackageWarningNoLine{StepRS}{Could not patch showhyphens; update microtype}}%
    }%
    \patchcmd{\MT@setup@expansion}
      {\MT@exp@cs\CheckCommand{showhyphens }}
      {\StepRS@patch@showhyphens}
      {}%
      {\PackageWarningNoLine{StepRS}{Could not install microtype compatibility repair}}%
  }
\fi
\makeatother
\usepackage{xcolor}
\usepackage{graphicx}
\usepackage{booktabs}
\usepackage{amsmath, amsfonts, amssymb}
\usepackage{amsthm}
\usepackage{mathtools}
\usepackage{algorithm}
\usepackage{algorithmic}
\usepackage{multirow}
\usepackage[inline, shortlabels]{enumitem}
\usepackage{xspace}
\usepackage{tabularx}
\usepackage{array}
\usepackage{xurl}
\usepackage{placeins}
\newcolumntype{Y}{>{\raggedright\arraybackslash}X}
\newcolumntype{L}[1]{>{\raggedright\arraybackslash}p{#1}}

\newcommand{\method}{StepRS-GRPO\xspace}

\newcommand{\KL}{D_{\mathrm{KL}}}
\newcommand{\clip}{\operatorname{clip}}

\usepackage[most]{tcolorbox}

\definecolor{lightblue}{HTML}{a8dadc}
\definecolor{green}{HTML}{76c893}
\definecolor{coralred}{HTML}{f28482}
\definecolor{purple}{HTML}{7b8cde}

\usepackage{color, colortbl}
\usepackage[dvipsnames]{xcolor}
\usepackage{pbox}
\usepackage{tabularx}
\usepackage{float}
\usepackage{wrapfig,lipsum}
\usepackage{graphicx}
\usepackage{pgfplots}
\pgfplotsset{compat=1.18}
\usepackage[T1]{fontenc}
\usepackage{inconsolata} 
\usetikzlibrary{arrows,shapes,automata,backgrounds,petri}
\usepackage{booktabs} %

\newcommand{\problemdivider}[1][gray!45]{%
  \par\medskip
  \noindent{\color{#1}\rule{\linewidth}{0.6pt}}%
  \par\medskip
}

\titlecontents{appsection}
  [15pt]
  {\addvspace{8pt}\normalfont\bfseries}
  {\contentslabel[\thecontentslabel\hfill]{15pt}}
  {\hspace*{-15pt}}
  {\hfill\contentspage}

\titlecontents{appsubsection}
  [39pt]
  {\addvspace{2pt}\normalfont}
  {\contentslabel[\thecontentslabel\hfill]{24pt}}
  {\hspace*{-24pt}}
  {\titlerule*[8pt]{.}\contentspage}

\renewcommand{\arraystretch}{1.15}

\title{Exploring More, Reasoning Better: \\
Stepwise Risk-Sensitive GRPO for Diffusion Language Models}

\author{%
  Yue Yu\textsuperscript{1,}\thanks{%
    Contributed equally to this work.
    Emails: \texttt{yyu3@iu.edu} and
    \texttt{bowen.zuo@email.ucr.edu}.}%
  \quad Bowen Zuo\textsuperscript{2,}\footnotemark[1]%
  \quad David Crandall\textsuperscript{1}%
  \quad Yinglun Zhu\textsuperscript{2}%
  \quad Dongruo Zhou\textsuperscript{1}\\[0.6em]
  {\normalsize\textsuperscript{1}\,Indiana University\quad
  \textsuperscript{2}\,University of California, Riverside}
}

\begin{document}
\date{}
\maketitle

\begin{abstract}
Diffusion large language models (dLLMs) generate text by denoising a sequence or successive blocks, allowing several tokens to be revealed in parallel. Reinforcement learning with verifiable rewards (RLVR) reuses terminal feedback across these decisions, even as their conditioning context changes. We propose stepwise risk-sensitive GRPO (\method), which modifies the GRPO advantage through a risk coefficient that depends on the denoising state. For binary rewards, we show that this transformation is exactly a prompt- and state-dependent rescaling of centered outcome advantages. A capability-based calibration suggests a coefficient scale, while endpoint and interpolation ablations guide schedule selection. Across multiple dLLM backbones and mathematical reasoning benchmarks, \method improves both pass@$1$ accuracy and pass@$k$ coverage over centered GRPO, while increasing answer diversity. In our ablation studies, mass-matched controls support state-dependent allocation over the tested prompt-only and reversed-schedule alternatives, and the gains persist after matching
the root mean square (RMS) of the advantages to that of centered GRPO. Reasoning-trace diagnostics further show that the diversity gains from \method extend beyond final-answer strings.

\end{abstract}

\section{Introduction}
\label{sec:intro}

Diffusion large language models (dLLMs)~\citep{nie2025llada,sahoo2024mdlm} change the structure of language-model generation. Rather than generating tokens strictly from left to right, a dLLM starts from a masked sequence and gradually denoises it through $T$ reverse steps, where each step may unmask multiple positions using bidirectional context. Recent work has adapted supervised post-training to this masked-denoising paradigm, either by training instruction-following masked dLLMs directly~\citep{zhao2025d1,nie2025llada} or by converting strong autoregressive backbones into bidirectional masked dLLMs through supervised distillation~\citep{wang2025tracerl}.

Reinforcement learning with verifiable rewards (RLVR) improves reasoning by optimizing outcome rewards, with methods such as Group Relative Policy Optimization (GRPO)~\citep{shao2024deepseekmath} now adapted to dLLMs~\citep{zhu2025dirl}. Existing dLLM RL studies have largely emphasized single-sample reasoning accuracy, measured by pass@$1$~\citep{zhao2025d1,stabledrl2026}. How post-training affects solution coverage under repeated sampling, measured by pass@$k$, remains less explored. A higher pass@$1$ alone does not establish better pass@$k$: improved accuracy can accompany a more concentrated output distribution. While pass@$k$-oriented optimization has been studied in RLVR~\citep{walder2026pkpo}, dLLMs offer a further design dimension: the context available to the policy changes during denoising, yet a terminal reward can supply the same advantage across these states. This raises the question of whether state-dependent advantage weighting can improve both accuracy and repeated-sampling coverage:

\begin{center}
    \emph{Can we design RLVR methods for dLLMs that improve reasoning accuracy while preserving answer coverage?}
\end{center}

We address this question with \emph{stepwise risk-sensitive GRPO} (\method). Our contributions are threefold:

\begin{itemize}[leftmargin=1.5em,itemsep=0.2em,topsep=0.2em]
\item \textbf{State-dependent advantage weighting.} Building on RS-GRPO~\citep{jiang2025rsgrpo}, we make the risk coefficient depend on the fraction of masked tokens in the denoising state, assigning larger coefficients to highly masked states and smaller ones as tokens are revealed. The method uses terminal rewards alone, without additional intermediate-state completions or process rewards. For binary rewards, we establish an exact finite-group factorization as prompt- and state-dependent rescaling of centered outcome advantages.
\item \textbf{Accuracy and repeated-sampling coverage.} We evaluate \method on SDAR-1.7B, SDAR-4B, and a Qwen3-0.6B masked-dLLM case study using mathematical reasoning benchmarks. Across the evaluated backbones and benchmarks, it improves mean pass@$k$ over centered GRPO at every tested sampling budget, including pass@$1$. These gains accompany higher unique-answer ratios; answer-removed trace diagnostics on SDAR-4B further show increased variation beyond final-answer strings.
\item \textbf{The value of scheduling.} Comparisons with strong constant-risk baselines show that state-dependent scheduling improves coverage beyond constant-risk weighting in the tested settings. Prompt-only and reversed-schedule controls examine the allocation of weights across states, while RMS matching shows that gains persist after matching overall advantage magnitude. We also develop a capability-guided starting scale for coefficient selection and compare endpoints and interpolation shapes, supporting linear half-decay as a practical default.
\end{itemize}

\section{Related Work}
\label{sec:related}

\paragraph{Diffusion modeling and efficient generation.}
Advances in diffusion language modeling address how to construct generators and reduce sampling cost. Trajectory self-distillation trains few-step students to match full-step teacher trajectories~\citep{zhang2026t3d}, while S2D2 uses a block-diffusion model's autoregressive mode for training-free self-speculative verification~\citep{han2026s2d2}. Uno augments an autoregressive model with diffusion parameters to sample in parallel while preserving its autoregressive target distribution~\citep{sahoo2026uno}. Uniform-diffusion analysis further separates denoiser parameterization, training objectives, and reverse sampling dynamics~\citep{gourevitch2026uniform}. These works study generator and sampler design, whereas StepRS studies outcome-advantage weighting within a selected generation and training framework.

\paragraph{Policy optimization for diffusion models.}
dLLM RL methods adapt policy optimization to non-autoregressive denoising. d1 combines masked supervised fine-tuning with diffu-GRPO~\citep{zhao2025d1}; TraceRL optimizes inference traces~\citep{wang2025tracerl}; StableDRL uses unconditional clipping and self-normalization~\citep{stabledrl2026}; and dTRPO reduces trajectory probability-estimation cost~\citep{zhang2026dtrpo}. Likelihood-estimation approaches include GDPO, ESPO, AGRPO, SPG, and BGPO~\citep{rojas2025gdpo,ou2025espo,zhan2025agrpo,wang2025spg,lin2025bgpo}. The ratio-free wd1 objective uses weighted log-likelihood, and wd1++ extends it to denoising-stepwise optimization~\citep{tang2025wd1}. DMPO matches a reward-tilted target through weighted denoising cross-entropy~\citep{zhu2026dmpo}. Recent GRPO theory separates the effects of reward normalization and reference- versus previous-policy KL regularization~\citep{mroueh2026grpo}. These works inform the construction and interpretation of the training objective; StepRS studies a state-dependent entropic advantage transformation, which we evaluate within TraceRL and StableDRL (Appendix~\ref{app:native_updates}).

Rollout and masking choices also determine the states used for policy optimization. JustGRPO uses left-to-right RL rollouts while retaining parallel decoding at inference~\citep{ni2026flexibility}. SLIM-RL combines confidence-budgeted rollout decoding with trace-free random-masking optimization, sequence-level importance sampling, and masking-level quadrature~\citep{zhao2026slimrl}. Its per-block mask schedule changes the construction of training states, and its risk parameter budgets confidence-based token commits. StepRS instead changes the entropic outcome advantage supplied at a given denoising state, addressing a complementary aspect of post-training to rollout generation and random-masking optimization.

\paragraph{Regularization and distribution preservation.}
Distribution-preserving approaches seek to limit the loss of diversity during reward optimization. Diversity-aware RLVR includes entropy analyses~\citep{cui2025entropy}, divergence-based methods~\citep{li2025divergence}, explicit diversity rewards~\citep{li2025darling}, and adaptive entropy targets~\citep{wang2025aepo}. Entropy-regularized RL also uses temperature-like coefficients to balance reward optimization and stochasticity~\citep{haarnoja2018sac,haarnoja2018sacapps}. Concurrent work, IDRF, combines trajectory-based reward optimization for few-step masked generators with inverse-distillation regularization~\citep{gromadskii2026idrf}, studying reward alignment and preservation of reference-distribution properties across text, image, and DNA generation. These approaches motivate studying distributional effects alongside reward gains; StepRS investigates state-dependent outcome-advantage weighting without introducing an additional diversity reward or distribution-preservation regularizer.

\section{Method: From GRPO to StepRS-GRPO}
\label{sec:method}
\method makes the risk-sensitive advantage depend on the denoising
state. We first specify the GRPO objective, then define the
state-dependent transformation and interpret it under binary rewards.

\subsection{Setup and notation}
\label{sec:prelim:dllm}
\paragraph{Diffusion language models.}
For a prompt $q$, a dLLM generates a response through reverse states
$s_T,s_{T-1},\ldots,s_0$. Each state contains the partially revealed
text and decoder information, including the active block when decoding
blockwise~\citep{sahoo2024mdlm,jetlm2025sdar}. Let
$\operatorname{out}(s_0)=x_0$ extract the completed text. Conditional
on the initialized state $s_T$, the output distribution is
\[
\pi_\theta(x_0\mid q,s_T)
=\sum_{\substack{s_0,\ldots,s_{T-1}:\\
                 \operatorname{out}(s_0)=x_0}}
\prod_{t=1}^{T}\pi_\theta(s_{t-1}\mid s_t,q,t),
\]
where $\theta$ denotes model parameters. Early termination is padded
with absorbing transitions to a common horizon $T$. The sum
marginalizes all denoising paths ending in the same text; the
conditional token probabilities used below are distinct from this
output marginal. Appendix~\ref{app:denoising_trajectories} gives
further details.

\paragraph{GRPO on denoising decisions.}
A frozen policy $\pi_{\mathrm{old}}$ samples $G$ responses $x_{qi,0}$ with binary outcome rewards
$r_{qi}\in\{0,1\}$. We use mean-centered GRPO advantages, without
reward-standard-deviation normalization \citep{shao2024deepseekmath}:
\[
\widehat p_q=\frac1G\sum_{i=1}^{G}r_{qi},\qquad
A^0_{qi}=r_{qi}-\widehat p_q.
\]
Let $j=(i,t,\ell)\in\mathcal J_q$ index a token revealed in response
$i$ at position $\ell$ from the pre-reveal state $s_{i,t}$; $i(j)$
identifies its response. Centered GRPO assigns
$a_{qj}=A^0_{q,i(j)}$ to every such decision.

For an advantage array $A$, the classical clipped loss is
\begin{align}
\mathcal L_{\mathrm{clip}}(\theta;A)
={}&-\frac1{|\mathcal B|}\sum_{q\in\mathcal B}
\sum_{j\in\mathcal J_q}v_{qj}
\min\!\left\{\rho_{qj}A_{qj},
\clip(\rho_{qj},1-\epsilon,1+\epsilon)A_{qj}\right\}
\nonumber\\
&+\lambda_{\mathrm{KL}}\mathcal K_{\mathrm{ref}}(\theta).
\label{eq:grpo_loss}
\end{align}
Here $\mathcal B$ is the prompt minibatch, $\rho_{qj}$ is the
current-to-old conditional token-probability ratio at the same
recorded state, $\epsilon>0$ is the ratio clip, and $v_{qj}>0$ are
fixed response/token reduction weights. The optional term
$\lambda_{\mathrm{KL}}\mathcal K_{\mathrm{ref}}$ penalizes categorical
KL divergence to a fixed reference policy. Thus centered GRPO is
$\mathcal L_{\mathrm{clip}}(\theta;a)$
\citep{schulman2017ppo,shao2024deepseekmath}.
Appendices~\ref{app:dllm_families} and~\ref{app:direct_implementation}
give the diffusion background and full objective definitions.

\subsection{State-dependent risk-sensitive advantages}
\label{sec:method:grpo}
\paragraph{A coordinate for the denoising state.}
Let $\mathcal I_{qj}$ contain the valid response positions in the
active decoding block and let $\widetilde x_{qj}$ be the text in
$s_{i,t}$. We measure local maskedness by
\begin{equation}
u_{qj}=\frac{1}{|\mathcal I_{qj}|}
\sum_{p\in\mathcal I_{qj}}
\mathbf 1\{\widetilde x_{qj,p}=\texttt{[MASK]}\}.
\label{eq:state_mask_fraction}
\end{equation}
Prompt, padding, and future-block positions are excluded; a partial
block uses its actual valid size. For response-wide decoding, the
block is the whole response canvas. The coordinate restarts at
$u=1$ for each fully masked block and decreases as positions are
revealed.

\paragraph{The advantage transformation.}
For a constant risk coefficient $\beta>0$, RS-GRPO
\citep{jiang2025rsgrpo} replaces the centered advantage with
\begin{equation}
A_{qi}(\beta)
=\frac1\beta\left[
\frac{\exp(\beta r_{qi})}
{G^{-1}\sum_{h=1}^{G}\exp(\beta r_{qh})}-1
\right],\qquad A_{qi}(0)=A^0_{qi}.
\label{eq:rs_adv_method}
\end{equation}
\method instead chooses $\beta$ from the current state through a
fixed map $b:[0,1]\to[0,\infty)$:
\begin{equation}
\beta_{qj}=b(u_{qj}),\qquad
A^{\mathrm{Step}}_{qj}=A_{q,i(j)}(\beta_{qj}),\qquad
B_{qj}=\clip(A^{\mathrm{Step}}_{qj},-C,C).
\label{eq:rs_adv_step}
\end{equation}
The cap $C\ge1$ limits advantage magnitude. For each decision $j$,
its coefficient $b(u_{qj})$ is applied to all $G$ rewards in the group
normalizer of Equation~\eqref{eq:rs_adv_method}. The resulting loss is
\begin{equation}
\mathcal L_{\mathrm{StepRS}}(\theta)
=\mathcal L_{\mathrm{clip}}(\theta;B),
\label{eq:steprs_loss}
\end{equation}
The state coordinates and advantages are detached during
optimization. Setting $b\equiv0$ recovers centered GRPO; a constant
$b>0$ recovers capped RS-GRPO. Only outcome rewards are required.

\paragraph{A simple schedule.}
We use linear interpolation with
$\beta_{\max}\ge\beta_{\min}\ge0$:
\begin{equation}
b(u)=\beta_{\min}+(\beta_{\max}-\beta_{\min})u,
\qquad \beta_{\min}=\gamma\beta_{\max},\quad 0\le\gamma\le1.
\label{eq:beta_schedule}
\end{equation}
This assigns the upper coefficient to fully masked blocks and
reduces it as context is revealed. Our half-decay choice sets
$\gamma=1/2$. The lower endpoint at $u=0$ is fully revealed and
contributes no masked-token loss, so the last active coefficient
need not equal $\beta_{\min}$. Endpoint values and schedule ablations
are reported in Section~\ref{sec:experiments}.

Algorithm~\ref{alg:step_rs_grpo} summarizes the method.
Appendix~\ref{app:direct_implementation} gives the objective and
optimization details. Experimental adaptations are described in
Appendix~\ref{app:native_updates}.

\begin{algorithm}[!htbp]
\caption{StepRS-GRPO}
\label{alg:step_rs_grpo}
\small
\begin{algorithmic}[1]
\REQUIRE Policy $\pi_\theta$, verifier $R$, group size $G$, schedule $b$, and advantage cap $C$.
\FOR{each rollout cycle}
    \STATE Freeze $\pi_{\mathrm{old}}\leftarrow\pi_\theta$ and sample a prompt batch $\mathcal B$.
    \STATE For each $q\in\mathcal B$, sample $G$ responses; record decisions $\mathcal J_q$ and mask fractions $u_{qj}$.
    \STATE Evaluate outcome rewards $r_{qi}=R(x_{qi,0},q)$.
    \FOR{each $q\in\mathcal B$ and $j\in\mathcal J_q$}
        \STATE Set $\beta_{qj}=b(u_{qj})$ and $A^{\mathrm{Step}}_{qj}=A_{q,i(j)}(\beta_{qj})$.
        \STATE Set $B_{qj}=\clip(A^{\mathrm{Step}}_{qj},-C,C)$ and detach it.
    \ENDFOR
    \STATE Update $\theta$ using $\mathcal L_{\mathrm{clip}}(\theta;B)$ with fixed state-dependent advantages (Equation~\eqref{eq:steprs_loss}).
\ENDFOR
\end{algorithmic}
\end{algorithm}

\subsection{Binary-reward interpretation and coefficient selection}
\label{sec:theory:tradeoff}
Under binary rewards, the transformation has an exact reweighting
interpretation.
\begin{lemma}[Finite-group binary-reward factorization]
\label{lem:variance_ratio}
For every realized binary-reward group and $\beta>0$,
\begin{equation}
A_{qi}(\beta)=c_\beta(\widehat p_q)A^0_{qi},\qquad
c_\beta(p)=\frac{e^\beta-1}{\beta(1-p+pe^\beta)},
\qquad 0\le p\le1.
\label{eq:cbeta_main}
\end{equation}
The identity extends to $\beta=0$ with $c_0(p)=1$.
\end{lemma}
The proof is in Appendix~\ref{app:theory_thm2}. Consequently,
$A^{\mathrm{Step}}_{qj}=c_{b(u_{qj})}(\widehat p_q)a_{qj}$ before
capping: StepRS preserves the outcome-advantage sign while changing
its weight across prompts and states. At a fixed coefficient,
$c_\beta(p)$ decreases with $p$, assigning larger multipliers to
groups with lower solve fractions. Increasing $\beta$ need not
increase magnitude at every solve rate. The experiments compare
this allocation with constant-risk, prompt-only, reversed-schedule,
and advantage-RMS-matched controls.

\paragraph{Capability-guided coefficient selection.}
Exponential tilting by $e^{\beta R}$ reduces failure-to-success odds
by a factor $e^{-\beta}$. Matching the fractional reduction
$1-e^{-\beta}$ to an initial-policy solve-rate proxy $\bar p$ suggests
\begin{equation}
\beta^\star(\bar p)=-\log(1-\bar p),\qquad 0\le\bar p<1.
\label{eq:beta_scaling}
\end{equation}
This is a starting scale for endpoint search, not an optimum derived
from the training loss. Selected endpoints remain fixed during a run.
Appendix~\ref{app:additional_theory} defines the calibration proxy
and gives the tilting argument. Neither the factorization nor this calibration
identifies the clipped update with an exact entropic trajectory
objective.
\FloatBarrier

\section{Experiments}
\label{sec:experiments}
We first describe the experimental setup, then report accuracy,
solution coverage, and output variation. Ablations examine the
contributions of state-dependent weighting and advantage scale,
followed by schedule sensitivity. Full setup details, additional
results, and extended ablations are given in
Appendices~\ref{app:experimental},~\ref{app:additional_results},
and~\ref{app:additional_ablations}, respectively.

\subsection{Experimental Setup}
\label{sec:experiments:setup}
\paragraph{Models and training.}
We conduct the main experiments using two block diffusion language models from the SDAR-Chat family~\citep{jetlm2025sdar}. \textbf{SDAR-1.7B-Chat} is a 1.7B-parameter block dLLM, while \textbf{SDAR-4B-Chat} is a 4B-parameter block dLLM from the same architectural family. Both models use the MATH training split restricted to difficulty levels 3--5 and the StableDRL framework~\citep{stabledrl2026}. We use a common training recipe within the SDAR family. Training runs for $160$ rollout cycles on NVIDIA H100 GPUs, with one old-policy refresh per cycle; we report the best evaluated checkpoint for each trained variant. For both models, we use $G=16$ rollouts per prompt and binary math-correctness rewards. An additional case study on masked dLLMs is provided in Appendix~\ref{app:exp_zerosixb}. Full hyperparameter details are reported in Appendix~\ref{app:experimental}.

\paragraph{Benchmarks and metrics.}
We evaluate mathematical reasoning on \textbf{MATH500}~\citep{hendrycks2021math}, \textbf{AMC23}~\citep{maa2023amc}, \textbf{AIME2024}~\citep{maa2024aime}, and \textbf{AIME2025}~\citep{maa2025aime}, covering both standard and competition-style math problems.

Accuracy and solution coverage are measured by pass@$k$, estimated
from $N=64$ responses per problem at sampling temperature $1.0$.
We also report unique-answer ratios and answer-removed trace
variation. Appendix~\ref{app:evaluation_protocol} defines the metrics,
answer verification, and response-budget rules.

\paragraph{Baselines and repeated evaluation.}
We compare the pre-RL model, centered GRPO, constant-risk RS-GRPO,
and \method under a common within-backbone trainer, optimizer,
verifier, and regularization setting
(Appendix~\ref{app:native_updates}).
All advantage-based variants apply the common cap $C=10$ before
any control-specific rescaling; C3 applies its RMS matching afterward.
The scheduled defaults are $[1.0\to0.5]$ for SDAR-1.7B,
$[1.5\to0.75]$ for SDAR-4B, and $[0.5\to0.25]$ for the
masked-model case study.
We repeat inference three times for each fixed checkpoint, using
different random seeds and evaluating
$k\in\{1,2,4,8,16,32,64\}$. These repeats quantify inference-sampling
variability, not variation across independent training runs.
Curves show means with capped error bars for one sample standard
deviation across fixed-checkpoint evaluations.
The mechanism controls are described in
Section~\ref{sec:experiments:ablations}.

\paragraph{Checkpoint and schedule selection.}
Checkpoint and risk-schedule selection was guided by inspection of the
overall pass@$k$ curves from evaluation, rather than a prespecified
single-$k$ selection score. We report the best evaluated checkpoint
for each trained variant, including centered GRPO, constant-risk
variants, StepRS, and the mechanism controls. For a given backbone, each evaluated policy
uses the same checkpoint across benchmarks, and the selected StepRS
schedule is also shared across benchmarks. Evaluation therefore informs
model selection as well as performance reporting; we do not treat these
scores as an untouched test-set assessment.

\subsection{Main Results}
\label{sec:experiments:main_results}
\label{sec:experiments:blockdllm}
\label{sec:experiments:onepointsevenb}
\label{sec:experiments:fourb}
\begin{figure}[t]
\centering
\includegraphics[width=\linewidth]{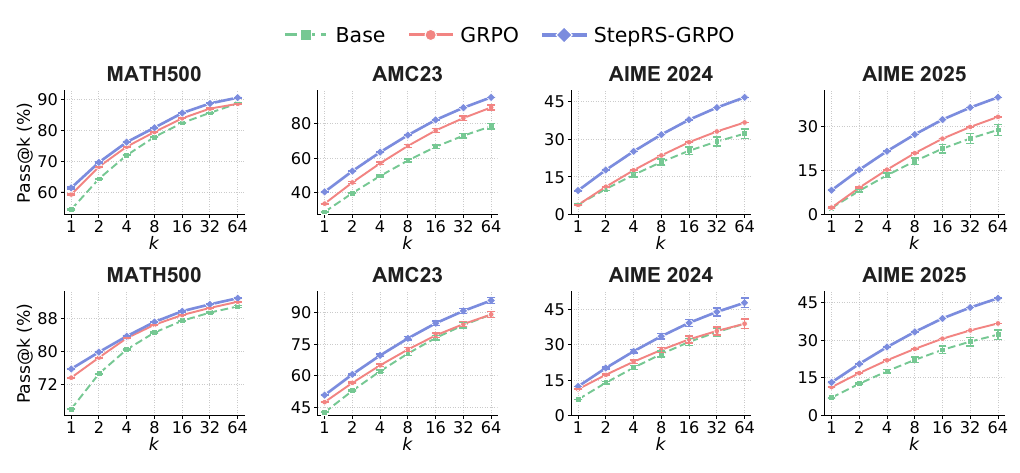}
\caption{Pass@$k$ on SDAR-1.7B-Chat (top row) and SDAR-4B-Chat
(bottom row) at the best evaluated checkpoints selected during $160$
rollout cycles (old-policy refreshes). Markers show means and capped error bars
show one sample standard deviation across three inference repeats,
with $64$ responses per problem in each repeat.
The selected StepRS schedules are $[1.0\to0.5]$ and
$[1.5\to0.75]$, respectively. Answer diversity is reported in
Figure~\ref{fig:unique_answer_all_backbones}.}
\label{fig:passk_block_dllm}
\end{figure}

Figure~\ref{fig:passk_block_dllm} compares Base, GRPO, and \method
on the two SDAR backbones. The selected schedules are
$[1.0\to0.5]$ for SDAR-1.7B and $[1.5\to0.75]$ for SDAR-4B.

\paragraph{Accuracy and solution coverage.}
We compare StepRS with centered GRPO to test whether state-dependent
weighting improves both single-sample accuracy and coverage under
repeated sampling. StepRS achieves higher mean pass@$k$ than centered GRPO at every
evaluated budget on all eight SDAR model--benchmark pairs
(Figure~\ref{fig:passk_block_dllm}).
The curves show how the comparison changes with the sampling budget.
On SDAR-1.7B, the pass@$1$ gains are particularly large on the AIME
benchmarks: $3.90\to9.60$ on AIME2024 and $2.40\to8.30$ on AIME2025.
The corresponding pass@$64$ gains are approximately $10.0$ and $6.7$
percentage points. On SDAR-4B, the AIME gains are approximately
$1.2$ and $1.9$ points at $k=1$, and $8.9$ and $10.0$ points at $k=64$.
Thus the improvements in these settings extend to the probability of
obtaining at least one verified solution under repeated sampling.

The masked-dLLM case study shows the same ordering over GRPO
across all seven budgets on GSM8K, MATH500, and AMC23
(Appendix~\ref{app:exp_zerosixb}).

\paragraph{Answer and reasoning-trace variation.}
\label{sec:trace_diagnostic}
We use output-variation diagnostics to complement verifier-based
accuracy and coverage. Unique-answer ratios summarize distinct
normalized final answers, including incorrect ones; they do not count
distinct correct solution methods. To examine variation beyond final
answers, we also measure the retained reasoning text.

The coverage gains accompany higher unique-answer ratios on every
SDAR model--benchmark pair. On MATH500, for example, $\bar U$
increases from $19.20$ to $28.00$ at 1.7B and from $12.27$ to
$18.60$ at 4B.
\begin{figure}[htbp]
    \centering
    \includegraphics[width=0.90\linewidth]{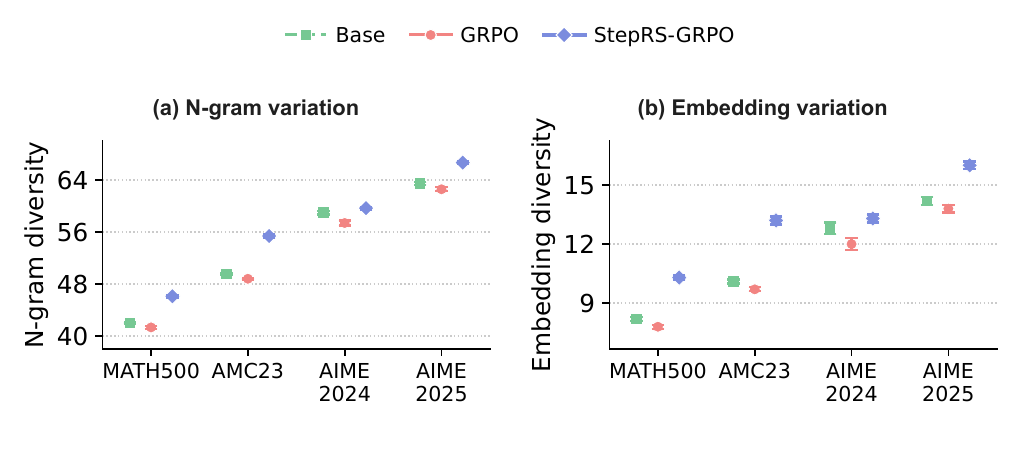}
    \caption{Answer-removed trace diversity on SDAR-4B:
    (a) $D_{\mathrm{ngram}}$; (b) $D_{\mathrm{emb}}$.
    Points and capped bars show means and sample standard deviations
    across three inference seeds, with $64$ responses per question
    per seed. Higher scores indicate greater variation; panel scales
    differ. Definitions are in
    Appendix~\ref{app:evaluation_protocol}.}
    \label{fig:trace_diversity}
\end{figure}

We remove the final boxed answer and its surrounding sentence before
measuring the remaining reasoning text. N-gram variation uses
$D_{\mathrm{ngram}}=100(1-\mathrm{Self\text{-}BLEU\text{-}4})$:
each trace is compared with the other traces for the same question
\citep{zhu2018texygen}. Embedding variation, $D_{\mathrm{emb}}$,
is $100$ times their mean pairwise cosine distance, using the
Sentence Transformers checkpoint \texttt{all-MiniLM-L6-v2}
\citep{reimers2019sentencebert,stAllMiniLML6v2}.
In Figure~\ref{fig:trace_diversity}, StepRS exceeds GRPO on all four
benchmarks under both measures, with gains of $2.3$--$6.6$ and
$1.3$--$3.5$ points, respectively. These all-trace diagnostics show
that the increased variation extends beyond final-answer strings;
Appendix~\ref{app:evaluation_protocol} defines both metrics in detail.
These diagnostics include incorrect traces and can reflect differences
in wording, length, or reasoning content. They therefore provide
evidence of textual variation alongside improved coverage, without
establishing a larger number of distinct correct reasoning strategies.

\FloatBarrier
\subsection{Ablation Studies}
\label{sec:experiments:ablations}
\begin{figure}[t]
\centering
\includegraphics[width=\linewidth]{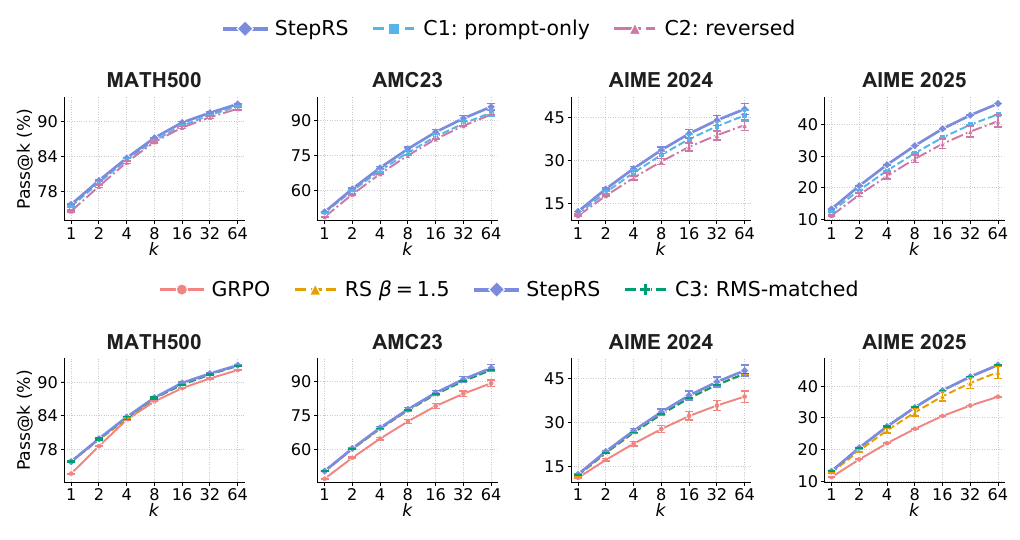}
\caption{SDAR-4B mechanism ablations.
Top: StepRS versus prompt-only mass matching (C1) and a
mass-normalized reversed schedule (C2).
Bottom: GRPO, constant-risk RS-GRPO ($\beta=1.5$), StepRS, and
advantage-RMS-matched StepRS (C3).
Markers show three-inference-seed means; capped error bars show
one sample standard deviation.
Controls are summarized in the text; formal definitions and matching
properties are given in Appendix~\ref{app:mechanism_controls}.}
\label{fig:ablation_mechanism_controls}
\end{figure}

We use SDAR-4B to distinguish the contributions of risk sensitivity,
denoising-state allocation, and overall advantage scale.
Figure~\ref{fig:ablation_mechanism_controls} compares all seven
sampling budgets under the common setup in
Section~\ref{sec:experiments:setup}. The controls test specified
alternatives to the reference weighting; their matching identities
hold on a fixed arm's sampled actions, while separately trained arms
can visit different trajectories. Results at different $k$ summarize
the same generated responses, rather than independent replications.

\paragraph{Scheduling versus constant risk.}
\label{sec:experiments:ablations_constant}
To test whether varying risk across denoising states adds value beyond
constant-risk weighting, we compare StepRS with constant-risk RS-GRPO.
The $\beta=1.5$ setting is a strong comparator.
\method matches or improves its mean pass@$k$ at every
benchmark--budget combination, with strict gains throughout AMC23
and both AIME benchmarks.
On these three benchmarks, the pass@$1$ gains are $0.20$, $0.40$,
and $0.50$ percentage points, and the pass@$64$ gains are $0.83$,
$1.10$, and $2.24$ points.
The full constant-risk sweep is shown in
Appendix~\ref{app:constant_risk_sweep}; the best constant can vary
with the benchmark and sampling budget. The scheduled-versus-constant
margins are smaller than several of the gains over centered GRPO.

\paragraph{State-dependent versus prompt-only weighting.}
To test whether prompt-dependent weighting alone accounts for the
gains, C1 removes state variation by replacing each multiplier with
its prompt-level loss-weighted mean. This preserves the reference
multiplier mass before advantage capping. StepRS has higher mean
pass@$k$ in all $28$ benchmark--budget comparisons, supporting
state-dependent allocation over this prompt-only alternative.

\paragraph{Proposed versus reversed coefficient assignment.}
To test the proposed assignment of coefficients to denoising states,
C2 evaluates the reversed coordinate $b(1-u)$ and matches the
reference multiplier mass before advantage capping. StepRS has
higher mean pass@$k$ in all $28$ comparisons, with pass@$64$ gains
of $5.58$ and $5.59$ points on AIME2024 and AIME2025. C2 changes
the active coefficient values as well as their order, so this
comparison does not isolate direction at a fixed coefficient distribution.

\paragraph{Matching overall advantage magnitude.}
To test whether the gains persist after matching overall advantage
magnitude, C3 rescales the capped StepRS advantages to match centered GRPO's
active-entry RMS, without recentering. Its pass@$1$ remains within
$0.24$ points of StepRS, and both variants exceed GRPO in all
comparisons. This matches advantage amplitude rather than the policy
gradient or optimizer update. Full control definitions, reduction
scopes, and matching properties are in
Appendix~\ref{app:mechanism_controls}.

\paragraph{Calibration and endpoint selection.}
\label{sec:experiments:schedules}
\label{sec:experiments:ablations_scaling}
To assess calibration as a starting scale for endpoint selection, we
compare the calibrated candidate with nearby half-decay schedules.
For SDAR-4B, the solve-rate proxy $\bar p\approx0.66$ gives the
calibrated candidate $[1.08\to0.54]$, whose pass@$1$ is within
$0.47$ points of the selected $[1.5\to0.75]$ schedule. This supports
calibration as a starting scale rather than a prospective selection
guarantee. Complete endpoint comparisons are in
Appendix~\ref{app:calibration_endpoints}.

\paragraph{Schedule shape at fixed endpoints.}
\label{sec:experiments:schedule_shape}
To assess sensitivity to interpolation independently of endpoint
choice, we fix the endpoints and compare linear, cosine, and two-stage
schedules. Linear and cosine interpolation differ
by at most $0.47$ points across the evaluated pass@$k$ values; both
exceed the tested two-stage schedule. We retain linear half-decay
as a simple default. Complete shape comparisons are in
Appendix~\ref{app:schedule_shape}; smaller-backbone results are in
Appendix~\ref{app:smaller_backbone_ablations}.
\FloatBarrier

\section{Conclusion and Limitations}
\label{sec:conclusion}
We introduced \method, a denoising-state-dependent extension of risk-sensitive GRPO for diffusion language models. Our experiments show improvements in mathematical accuracy, sampled-solution coverage, and output diversity. Ablation studies favor state-dependent allocation over the tested prompt-only and reversed-schedule alternatives, and the advantage-RMS-matched variant retains the gains over GRPO. Calibration and schedule-shape experiments further provide practical guidance for selecting effective coefficients.

The method is defined using a classical clipped GRPO objective;
the experiments apply its advantage transformation through native
trajectory-based and likelihood-proxy losses. Grouped contexts,
approximate caches, and proxy likelihoods need not preserve the direct
implementation's gradient. Our algebraic identities characterize the
advantage transformation, not the exactness of those adaptations.
The capability rule guides the selection of schedule endpoints, but does not establish an optimal schedule. Our evaluation also focuses on mathematical tasks with binary rewards, and the three inference seeds quantify sampling variability at fixed checkpoints rather than variability across training runs. Checkpoint and schedule selection also used the evaluation curves, so independently selected checkpoints and fresh held-out evaluation would strengthen the assessment. Additional training replications and evaluation under broader reward settings would help establish how consistently these gains hold beyond the settings studied here.

\appendix

\section*{Contents of the Appendix}
\startcontents[appendix]

\begingroup
  \hypersetup{linkcolor=black}

  \printcontents[appendix]{app}{1}[2]{%
    \setlength{\parskip}{0pt}%
    \contentsmargin{20pt}%
  }
\endgroup

\section{Diffusion Background}
\label{app:dllm_families}
This appendix distinguishes continuous diffusion time from the discrete
policy-loss index in Section~\ref{sec:prelim:dllm}. We suppress the
prompt conditioning only where it is not needed. Let $\mathcal V$ be
the clean-token vocabulary, $m\notin\mathcal V$ the mask token, and
$L$ the padded output length. The symbol
$\delta_a(z)=\mathbf 1\{z=a\}$ denotes point mass at token $a$;
$\mathbf 1\{E\}$ equals one when condition $E$ holds and zero otherwise.

\subsection{Masked diffusion}
\label{app:dllm_families:masked}
Use $\zeta\in[0,1]$ for continuous forward-diffusion time and
$\alpha(\zeta)$ for the masking probability, with
$\alpha(0)=0$, $\alpha(1)=1$, and $\alpha$ nondecreasing. Writing
$X_0=x_0\in\mathcal V^L$, the absorbing forward process has marginal
\begin{equation}
Q(X_\zeta^{(\ell)}=z\mid X_0=x_0)
=[1-\alpha(\zeta)]\delta_{x_0^{(\ell)}}(z)
+\alpha(\zeta)\delta_m(z),\qquad \ell=1,\ldots,L.
\label{eq:masked_forward}
\end{equation}
Masking is independent across positions and absorbing across time.
The expected masked fraction is $\alpha(\zeta)$; a realized state's
masked fraction need not equal this expectation.

Let $f_\theta^{(\ell)}(z\mid x_\zeta,\zeta,q)$ be the model's
clean-token probability at position $\ell$, extended by
$f_\theta^{(\ell)}(m\mid\cdot)=0$. For $0\le\xi<\zeta\le1$
with $\alpha(\zeta)>0$, the factorized absorbing reverse kernel is
\begin{equation}
P_\theta(X_\xi^{(\ell)}=z\mid x_\zeta,q)=
\begin{cases}
\delta_{x_\zeta^{(\ell)}}(z),&x_\zeta^{(\ell)}\ne m,\\[2pt]
\dfrac{\alpha(\xi)}{\alpha(\zeta)}\delta_m(z)
+\left[1-\dfrac{\alpha(\xi)}{\alpha(\zeta)}\right]
f_\theta^{(\ell)}(z\mid x_\zeta,\zeta,q),&x_\zeta^{(\ell)}=m.
\end{cases}
\label{eq:masked_reverse}
\end{equation}
The full kernel multiplies these terms over positions
\citep{austin2021d3pm,sahoo2024mdlm,shi2024simplified}.
This absorbing kernel is background; a confidence-based decoder that
selects reveal positions may implement a different transition rule.
The rollout policy and its likelihood estimator must describe that
implemented decoder.

Assume $\alpha$ is absolutely continuous and $\alpha(\zeta)>0$
for $0<\zeta\le1$. The corresponding weighted denoising objective is
\begin{equation}
\mathcal L_{\mathrm{MDLM}}(\theta)
=-\mathbb E\!\left[
\frac{\alpha'(\zeta)}{\alpha(\zeta)}
\sum_{\ell=1}^{L}\mathbf 1\{X_\zeta^{(\ell)}=m\}
\log f_\theta^{(\ell)}(X_0^{(\ell)}\mid X_\zeta,\zeta,q)
\right].
\label{eq:masked_elbo}
\end{equation}
The expectation is over training examples $(q,X_0)$,
$\zeta\sim\mathcal U(0,1)$, and $X_\zeta\sim Q(\cdot\mid X_0)$;
$\alpha'$ is the time derivative. Endpoint values are interpreted
by limits. Since $\alpha$ denotes masking rather than clean-token
retention, the weight is $\alpha'/\alpha$
\citep{sahoo2024mdlm,shi2024simplified}.

\subsection{Block diffusion}
\label{app:dllm_families:block}
Partition the padded sequence into $B_{\mathrm{blk}}$ blocks of length
$\ell_{\mathrm{blk}}$. The clean sequence is
$x_0=(x_0^{[1]},\ldots,x_0^{[B_{\mathrm{blk}}]})$, and
$x_0^{[<d]}$ is the prefix preceding block $d$. Block diffusion uses
\begin{equation}
\pi_\theta(x_0\mid q)
=\prod_{d=1}^{B_{\mathrm{blk}}}
\pi_\theta(x_0^{[d]}\mid x_0^{[<d]},q).
\label{eq:block_factorization}
\end{equation}
For a fixed within-block horizon $K_{\mathrm{blk}}$, pad any early
completion with deterministic identity transitions. Initialize
$x_{K_{\mathrm{blk}}}^{[d]}=m^{\ell_{\mathrm{blk}}}$ and marginalize
only the unfixed intermediate states:
\begin{equation}
\pi_\theta(x_0^{[d]}\mid x_0^{[<d]},q)
=\sum_{x_1^{[d]},\ldots,x_{K_{\mathrm{blk}}-1}^{[d]}}
\prod_{h=1}^{K_{\mathrm{blk}}}
P_\theta(x_{h-1}^{[d]}\mid x_h^{[d]},x_0^{[<d]},q,h).
\label{eq:block_within_diffusion}
\end{equation}
Attention is causal across blocks and bidirectional within the active
block~\citep{arriola2025bd3lm,jetlm2025sdar}. Prefix keys and values
can be cached, although attention still incurs a prefix-length-dependent
cost. Our SDAR experiments use $\ell_{\mathrm{blk}}=4$.

\paragraph{Relation to the risk coordinate.}
The trajectory index, continuous diffusion time $\zeta$, and expected
masking probability $\alpha(\zeta)$ are distinct from the realized
within-block mask fraction in Equation~\eqref{eq:state_mask_fraction}.
That coordinate restarts at $u=1$ for each fully masked block; dynamic
reveals can skip mask-count levels. A full-attention architecture can
also use a block-wise sampler, which restricts reveal order without
imposing a block-causal attention mask. The mapping from this state
coordinate to the experimental training inputs is specified in
Appendix~\ref{app:native_updates}.

\subsection{Denoising trajectories and conditional scores}
\label{app:denoising_trajectories}
For a prompt $q$, a dLLM follows reverse denoising states
$s_T,s_{T-1},\ldots,s_0$ with a common horizon $T$. Variable stopping
is represented by deterministic absorbing transitions after termination. A state contains the partially revealed text and the decoder information needed for the next transition. Let $\operatorname{out}(s_0)=x_0$ extract the final text. Conditional on the initialized state $s_T$, its distribution is
\[
\pi_\theta(x_0\mid q,s_T)
=\sum_{\substack{s_0,\ldots,s_{T-1}:\\
                 \operatorname{out}(s_0)=x_0}}
\prod_{t=1}^{T}\pi_\theta(s_{t-1}\mid s_t,q,t).
\]
Here $\theta$ denotes the model parameters. The sum includes every state trajectory ending in the same output, including any terminal decoder metadata. Deterministic initialization makes this also the distribution given $q$; otherwise, we average over $s_T$. For a trajectory $\tau$, we write $x_0(\tau)$ for this extracted text.

Let $p_\theta(v\mid s_{i,t},q,\ell)$ denote the categorical token
distribution scored at this decision. We use conditional token
probabilities in a policy surrogate; they are not the marginal
probability of the completed output, which sums over denoising paths.
In particular, evaluating original states without acceleration does
not remove the intractability of that sequence marginal.

\section{Additional Method Details and Analysis}
\label{app:additional_theory}
The empirical group fraction $\widehat p_q$, the population success
probability $p_\pi(q)$, and the model-level calibration average $\bar p$
serve different roles. The first enters the actual advantages; the
second describes a fixed rollout distribution; the third suggests a
coefficient scale before training. The identities below keep these
quantities separate.

\subsection{Proof of Lemma~\ref{lem:variance_ratio}}
\label{app:theory_thm2}
Fix a prompt $q$ and its realized rewards $r_{q1},\ldots,r_{qG}$.
Because each reward is either zero or one,
$e^{\beta r_{qi}}=1+r_{qi}(e^\beta-1)$. Averaging this equality
over the group gives the exact normalizer
\[
\widehat D_q(\beta)=\frac1G\sum_{i=1}^{G}e^{\beta r_{qi}}
=1+\widehat p_q(e^\beta-1).
\]
For $\beta>0$, substitute this expression into
Eq.~\eqref{eq:rs_adv_method} and collect the terms containing the
reward:
\begin{align}
A_{qi}(\beta)
&=\frac{1+r_{qi}(e^\beta-1)-[1+\widehat p_q(e^\beta-1)]}
{\beta[1+\widehat p_q(e^\beta-1)]}\nonumber\\
&=\frac{e^\beta-1}{\beta[1+\widehat p_q(e^\beta-1)]}
(r_{qi}-\widehat p_q)
=c_\beta(\widehat p_q)A^0_{qi}.
\label{eq:cbeta_proof}
\end{align}
This proves the finite-group factorization. If every reward is zero
or every reward is one, $r_{qi}=\widehat p_q$ for every response,
so all advantages are zero. At $\beta=0$, the expansion
$e^\beta-1=\beta+O(\beta^2)$ gives $c_\beta(p)\to1$.
Thus the identity extends continuously to the centered baseline.
These statements hold for each realized group, without an independence
assumption or a large-group limit.

\hfill$\square$

\paragraph{Interpretation of the multiplier.}
For a fixed coefficient $\beta>0$,
\[
\frac{\partial c_\beta(p)}{\partial p}
=-\frac{(e^\beta-1)^2}{\beta[1+p(e^\beta-1)]^2}<0.
\]
Thus, groups with lower observed solve fractions receive larger
multipliers at the same coefficient. This is a statement about relative
weighting across groups, not a guarantee that every multiplier
exceeds one. Moreover,
$c_\beta(p)=1+(\tfrac12-p)\beta+O(\beta^2)$ near zero;
increasing $\beta$ need not increase advantage magnitude uniformly
across solve rates. For a common coefficient, the group advantages
sum to zero. With action-dependent coefficients and token-reduction
weights, an action-weighted mean need not be zero, which is one reason
to distinguish RMS matching from standard-deviation matching.

\subsection{Population tilting and success odds}
\label{app:population_tilting}
Let $p=p_\pi(q)=\Pr_{\tau\sim\pi(\cdot\mid q)}[R(x_0(\tau),q)=1]$
be the success probability of a fixed rollout distribution, where
$x_0(\tau)$ is the completed response. Define its exponentially
tilted distribution by
\begin{equation}
\pi_\beta(\tau\mid q)
=\frac{\pi(\tau\mid q)\exp\{\beta R(x_0(\tau),q)\}}
{1-p+pe^\beta}.
\label{eq:tilted_rollout_distribution}
\end{equation}
Its success probability is $p_\beta=pe^\beta/(1-p+pe^\beta)$,
where $p=p_\pi(q)$. For $0<p<1$,
\begin{equation}
\log\frac{p_\beta}{1-p_\beta}
=\log\frac{p}{1-p}+\beta.
\label{eq:updated_success_prob}
\end{equation}
For this fixed distribution, let $p=p_\pi(q)$. Its normalizer is
$D_\beta=1-p+pe^\beta$. Summing the tilted probabilities separately
over successful and unsuccessful trajectories gives
\[
p_\beta=\frac{pe^\beta}{D_\beta},\qquad
1-p_\beta=\frac{1-p}{D_\beta}.
\]
When $0<p<1$, their ratio is $e^\beta p/(1-p)$; taking logarithms
proves Eq.~\eqref{eq:updated_success_prob}. At $p=0$ or $1$,
exponential tilting leaves the all-failure or all-success distribution
unchanged.

\paragraph{Connection to entropic utility.}
For a fixed prompt and binary reward, the population entropic utility
is $J_\beta(\pi;q)=\beta^{-1}\log[1+p_\pi(q)(e^\beta-1)]$.
Its derivative with respect to $p_\pi(q)$ is $c_\beta(p_\pi(q))$.
When differentiation under the rollout expectation is valid,
\[
\nabla_\theta J_\beta(\pi_\theta;q)
=c_\beta(p_{\pi_\theta}(q))\nabla_\theta p_{\pi_\theta}(q).
\]
This gives the population coefficient an outcome-reweighting
interpretation \citep{howard1972risk,jiang2025rsgrpo}.
The exact finite-group factorization above is a separate statement:
substituting the empirical fraction into a nonlinear normalizer does
not, by itself, establish an unbiased finite-group gradient estimator.

\subsection{Capability-guided calibration}
\label{app:capability}
Let $\pi_{\mathrm{init}}$ be the initial policy and
$\mathcal D_{\mathrm{cal}}$ a calibration distribution. Define
\[
\bar p=\mathbb E_{q\sim\mathcal D_{\mathrm{cal}}}
[p_{\pi_{\mathrm{init}}}(q)].
\]
This model-level quantity differs from the current training group's
$\widehat p_q$. Exponential tilting multiplies failure-to-success odds
by $e^{-\beta}$. Equating their fractional reduction
$g(\beta)=1-e^{-\beta}$ to the solve-rate proxy yields
\begin{equation}
\beta^\star(\bar p)=-\log(1-\bar p),\qquad 0\le\bar p<1.
\label{eq:beta_star_app}
\end{equation}
The rule suggests an interpretable search scale; it is not an
optimizer-derived optimum or upper bound. Empirical calibration and
endpoint comparisons are reported in
Appendix~\ref{app:calibration_endpoints}.

\subsection{Scope of denoising-state scheduling}
\label{app:step_schedule}
For each action $j$, StepRS applies the scalar group transformation
at $\beta_{qj}=b(u_{qj})$. All rewards in that evaluation of the group
normalizer use this same coefficient. Applying a different coefficient
to each reward inside one denominator would define another
transformation and would not give Equation~\eqref{eq:cbeta_main}.
The schedule therefore has a precise interpretation as the
advantage array substituted into the clipped objective in
Equation~\eqref{eq:steprs_loss}. The same algebra applies when that
array is supplied to an experimental native surrogate, but does not
make the two losses equivalent. Its direction and interpolation are
assessed by the controls and fixed-endpoint ablation, rather than
inferred from a fixed-coefficient trajectory objective.

\subsection{Classical objective and direct implementation}
\label{app:direct_implementation}
For target token $x_{qi,\ell}$, the conditional ratio in
Equation~\eqref{eq:grpo_loss} is
\begin{equation}
\rho_{qj}(\theta)=
\frac{p_\theta(x_{qi,\ell}\mid s_{i,t},q,\ell)}
     {p_{\mathrm{old}}(x_{qi,\ell}\mid s_{i,t},q,\ell)}.
\label{eq:conditional_ratio}
\end{equation}
The optional regularizer uses a fixed reference policy
$\pi_{\mathrm{ref}}$, typically initialized from the starting model:
\[
\mathcal K_{\mathrm{ref}}(\theta)=
\frac1{|\mathcal B|}\sum_{q\in\mathcal B}\sum_{j\in\mathcal J_q}v_{qj}
\KL\!\left(p_\theta(\cdot\mid s_{i,t},q,\ell)
\,\|\,p_{\mathrm{ref}}(\cdot\mid s_{i,t},q,\ell)\right).
\]
Its coefficient satisfies $\lambda_{\mathrm{KL}}\ge0$; zero omits it.
The reference remains fixed across training, whereas
$\pi_{\mathrm{old}}$ is refreshed to collect new trajectories.
States, target tokens, reduction weights, and supplied advantages
are held fixed during each loss evaluation.

A direct implementation records the original pre-reveal states and
evaluates their conditional token probabilities in separate forward
passes. It computes $u_{qj}$ and the corresponding advantage before
the fixed loss reduction. Neither trace shrinkage nor a KV-cache
approximation is required to define this algorithm. In the experiments,
trajectory grouping, likelihood proxies, and native surrogate choices
provide computationally practical adaptations of the same advantage
transformation; they need not reproduce every term of
Equation~\eqref{eq:grpo_loss} (Appendix~\ref{app:native_updates}).

Old-policy conditional probabilities are cached on the recorded
states before inner optimization. Each inner step evaluates current
probabilities on a minibatch of these decisions and updates the policy
using Equation~\eqref{eq:steprs_loss}. The old policy, recorded states,
rewards, and advantage labels remain fixed throughout the inner
optimization steps of a rollout cycle.
An old-policy refresh counts an outer cycle, not necessarily one
optimizer update; the fixed reference, when used, is not refreshed.
\FloatBarrier

\section{Experimental Setup Details}
\label{app:experimental}
This appendix collects configuration, native update rules, evaluation
metrics, and prompts for the experiments in
Section~\ref{sec:experiments:setup}.

\subsection{Decoding and training configuration}
\label{app:training_configuration}
Both training and evaluation generate responses using the decoding
configuration in Table~\ref{tab:shared_decoding}. Table~\ref{tab:training_hyperparams}
reports training data, sampling scale, response limits, optimization,
and hardware; Appendix~\ref{app:native_updates} specifies the native
update interface and implementation details. Evaluation sampling and benchmark-specific response
budgets are specified separately in Appendix~\ref{app:evaluation_protocol}.
Unless an experiment-specific value is stated, unmodified parameters
follow the corresponding upstream configuration: the TraceRL
full-attention path for Qwen3-MDLM and the StableDRL block-diffusion
configuration for SDAR \citep{wang2025tracerl,stabledrl2026}.
In particular, training sampling temperatures are $0.8$ and $1.0$,
respectively, and SDAR's training confidence threshold is $0.9$.
These phase-specific values are listed in
Table~\ref{tab:training_hyperparams}; evaluation temperature remains
$1.0$ and SDAR's evaluation threshold remains $0.85$.
The Qwen3-MDLM static sampler does not use a confidence threshold;
its upstream sampling path adds no top-$k$ or top-$p$ filtering.
The initial checkpoints are
\path{dllm-hub/Qwen3-0.6B-diffusion-mdlm-v0.1},
\path{JetLM/SDAR-1.7B-Chat}, and \path{JetLM/SDAR-4B-Chat}.

\begin{table}[htbp]
\centering
\caption{Selected schedules and evaluated constant-risk coefficients.}
\label{tab:closed_form_recipe}
\small
\begin{tabular}{@{}lll@{}}
\toprule
Backbone & Default StepRS schedule & Constant-risk sweep \\
\midrule
Qwen3-0.6B-MDLM & $[0.5\to0.25]$ & $\{0.25,0.5,1.0\}$ \\
SDAR-1.7B-Chat & $[1.0\to0.5]$ & $\{0.25,0.5,1.0\}$ \\
SDAR-4B-Chat & $[1.5\to0.75]$ & $\{0.5,1.0,1.5,2.0\}$ \\
\bottomrule
\end{tabular}
\end{table}

\begin{table}[!htbp]
\centering
\caption{Decoding configuration shared by training and evaluation.
Engines, cache settings, and decoding block lengths follow the respective
upstream implementations. Denoising budgets are expressed per block;
response limits and sampling scales are specified separately.}
\label{tab:shared_decoding}
\small
\setlength{\tabcolsep}{4pt}
\renewcommand{\arraystretch}{1.18}
\begin{tabularx}{\linewidth}{@{}>{\raggedright\arraybackslash}p{0.26\linewidth}YYY@{}}
\toprule
\textbf{Setting}
& \textbf{Qwen3-0.6B-MDLM}
& \textbf{SDAR-1.7B-Chat}
& \textbf{SDAR-4B-Chat} \\
\midrule
Model attention
& Full bidirectional
& \multicolumn{2}{c}{Block-causal; bidirectional within block} \\
Decoding block length
& $32$
& \multicolumn{2}{c}{$4$} \\
Reveal rule
& Static within block
& \multicolumn{2}{c}{Dynamic within each block} \\
Denoising steps per block
& $16$
& \multicolumn{2}{c}{At most $4$ steps per block} \\
Generation backend
& Fast-dLLM~\citep{wu2025fastdllm}
& \multicolumn{2}{c}{JetEngine~\citep{labman2025jetengine}} \\
KV-cache configuration
& Prefix; block-wise refresh
& \multicolumn{2}{c}{Reuse completed-block KV} \\
Further horizon (tokens)
& $128$
& \multicolumn{2}{c}{Not applicable} \\
\bottomrule
\end{tabularx}
\end{table}

\paragraph{Qwen3-MDLM decoding.}
The model has full bidirectional attention, while the sampler reveals
successive $32$-token blocks from left to right. We retain TraceRL's
Fast-dLLM-based prefix-cache sampler with a further horizon of $128$
tokens~\citep{wu2025fastdllm,wang2025tracerl}. During training, with a $512$-token response canvas,
there are $16$ decoding blocks. The configured $256$ total denoising
steps are divided evenly into $16$ steps per block; static sampling
commits two confidence-ranked positions within the current block per
step. Previously revealed token identities remain fixed. At the start
of each block, a full-sequence forward pass refreshes the prefix KV
cache. Subsequent steps reuse that cache and recompute the current
block together with up to $128$ following positions, while permitting
reveals only within the current block. Thus, full attention describes
the model architecture, not a response-wide reveal schedule.

\paragraph{SDAR decoding.}
SDAR uses a custom JetEngine~\citep{labman2025jetengine} rollout
implementation, as described by StableDRL~\citep{stabledrl2026}. It generates successive four-token
blocks from left to right, using confidence-based dynamic unmasking
for at most four denoising steps per block. The number of positions
committed in a step can vary, and a block can finish before exhausting
its step budget. Block-causal attention permits reuse of completed
blocks' KV without refreshing them when later blocks are generated.
Future blocks do not participate in the current block's attention.
The total number of rollout transitions depends on the generated
length and reveal pattern, rather than a fixed $256$-step response
horizon.

\begin{table}[!htbp]
\centering
\caption{Training configuration: data, response sampling, optimization,
and hardware. Response-token caps exclude the prompt. Structural
decoding settings are shared with evaluation
(Table~\ref{tab:shared_decoding}); phase-specific sampling parameters
are listed below. The KL specification is given in
Appendix~\ref{app:native_updates}.}
\label{tab:training_hyperparams}
\small
\setlength{\tabcolsep}{4pt}
\renewcommand{\arraystretch}{1.18}
\begin{tabularx}{\linewidth}{@{}>{\raggedright\arraybackslash}p{0.26\linewidth}YYY@{}}
\toprule
\textbf{Setting}
& \textbf{Qwen3-0.6B-MDLM}
& \textbf{SDAR-1.7B-Chat}
& \textbf{SDAR-4B-Chat} \\
\midrule
Training data
& GSM8K train
& \multicolumn{2}{c}{MATH train, levels 3--5} \\
Training responses per prompt ($G$)
& \multicolumn{3}{c}{16} \\
Training batch (prompts $\times$ responses)
& $64\times16=1024$
& \multicolumn{2}{c}{$32\times16=512$} \\
Training response-token cap
& 512
& \multicolumn{2}{c}{1024} \\
Reward
& \multicolumn{3}{c}{Binary final-answer correctness} \\
Trainer framework
& TraceRL~\citep{wang2025tracerl}
& \multicolumn{2}{c}{StableDRL~\citep{stabledrl2026}} \\
Trace shrinkage factor
& $8$ (original risk labels retained)
& \multicolumn{2}{c}{Not applicable} \\
Optimizer
& \multicolumn{3}{c}{AdamW} \\
Learning rate
& $5\times10^{-6}$
& \multicolumn{2}{c}{$1\times10^{-6}$} \\
Gradient accumulation steps
& 1
& \multicolumn{2}{c}{32} \\
Old-policy refresh cycles
& 100
& \multicolumn{2}{c}{160} \\
DeepSpeed
& ZeRO-3, CPU offload
& \multicolumn{2}{c}{ZeRO-3, no CPU offload} \\
Mixed precision
& \multicolumn{3}{c}{bf16} \\
GPUs
& $8\times$ L40S (48GB)
& \multicolumn{2}{c}{$8\times$ H100 (95GB)} \\
Additional KL coefficient
& $0.01$
& \multicolumn{2}{c}{$0$} \\
KL target
& Current rollout old policy
& \multicolumn{2}{c}{None} \\
KL estimator
& k3
& \multicolumn{2}{c}{Not applicable} \\
Advantage cap ($C$)
& \multicolumn{3}{c}{10.0} \\
Training sampling temperature
& $0.8$
& \multicolumn{2}{c}{$1.0$} \\
Training confidence threshold
& Not applicable (static)
& \multicolumn{2}{c}{$0.9$} \\
\bottomrule
\end{tabularx}
\end{table}
\FloatBarrier

\subsection{Native policy updates and regularization}
\label{app:native_updates}
Original run configurations were not retained. The setup description
combines the reported settings, author recollection, and adopted
upstream conventions; it is a reconstruction rather than an
independently verified record of the original runs.

\paragraph{Native-loss interface.}
We extend the decision index $j$ to the native training samples.
For TraceRL, its risk-label state is the original pre-reveal rollout
state, even if the forward input is grouped. For StableDRL, it is the
sampled corrupted target block; Equation~\eqref{eq:state_mask_fraction}
uses that block's realized mask count. For these samples, write
\begin{equation}
\mathcal L_{\mathrm{native}}(\theta;A)
=-\frac1{|\mathcal B|}\sum_{q\in\mathcal B}
\sum_{j\in\mathcal J_q}v_{qj}
\Psi^{\mathrm{native}}_{qj}(\theta;A)
+\mathcal R_{\mathrm{native}}(\theta).
\label{eq:native_loss}
\end{equation}
The positive weights $v_{qj}$ describe the platform's fixed loss
reduction. The contribution $\Psi^{\mathrm{native}}$ can depend on the
whole advantage array through native normalization. For each platform,
the centered baseline supplies $a$, while StepRS supplies $B$ from
Equation~\eqref{eq:rs_adv_step}; state labels, advantages, and fixed
reduction weights are detached during differentiation. This describes
an adapter for the advantage transformation, rather than asserting
that every platform optimizes Equation~\eqref{eq:grpo_loss}.

\paragraph{TraceRL surrogate and old-policy penalty.}
The Qwen3-MDLM path uses a PPO-style clipped conditional-token
surrogate on reconstructed trace inputs, with ratio clip $\epsilon=0.2$. Within each grouped training
row, active-token reward contributions are averaged, followed by a
minibatch average. For the configured penalty, old-policy token
log-probabilities are cached on the same inputs before policy updates.
Writing $d_j=\log p_\theta(x_j)-\log p_{\mathrm{old}}(x_j)$, the
k3 sample penalty is
\begin{equation}
k_{3,j}=\exp(-d_j)-1+d_j.
\label{eq:old_policy_k3}
\end{equation}
Following TraceRL, $\mathcal R_{\mathrm{native}}$ is $0.01$ times
the minibatch mean of the sum of $k_{3,j}$ over active target tokens
in each row~\citep{wang2025tracerl}. Unlike the reward term, this
penalty is not divided by each row's active-token count. Its target
is the policy frozen for the current rollout/update cycle, which is
refreshed for the next cycle, rather than a permanently frozen initial
reference. Thus it differs from the optional reference KL in the
classical method definition. The penalty coefficient is separate from
the StepRS risk coefficient $\beta$.

\paragraph{StableDRL surrogate and regularization choice.}
For SDAR, we insert state-dependent advantages into the target-block
contributions of the native StableDRL loss. The remaining settings
follow StableDRL~\citep{stabledrl2026}.
The implementation specification sets $\mathcal R_{\mathrm{native}}=0$:
there is no additional KL penalty. A behavior-policy proxy used in an
importance ratio is not a fixed-reference regularizer. Estimator
choices such as an ELBO/EUBO mixture and their sampling parameters
belong to the native loss configuration; they are not determined by
the rollout denoising budget or by the StepRS risk coefficient.

\paragraph{State labels and computational approximations.}
The rollout decoding budget is distinct from the construction of
policy-loss samples. The TraceRL path constructs training states from
recorded reveal traces~\citep{wang2025tracerl}. For Qwen3-MDLM,
we use shrinkage factor $s=8$: eight neighboring reveal rounds share
one training input, while each target token retains its original
pre-reveal mask fraction $u_{qj}$, coefficient $b(u_{qj})$, and
advantage. A full $32$-token block therefore yields two grouped
training inputs, each supervising $16$ target tokens, while retaining
all $16$ original risk stages. Token-level advantages are applied
before loss reduction. Shrinkage reduces forward computation by
approximating the conditioning context; it does not preserve the exact
unshrunk log-probabilities or gradients. The StableDRL path
estimates likelihood proxies using sampled corruptions with the
block-conditioned attention structure~\citep{stabledrl2026}.
Each sampled corruption retains its target-block contributions. We
compute $u_{qj}$ from the remaining masks in each target block and
apply the corresponding advantage before the trainer's final loss
reduction. If one corrupted input includes several target blocks,
their coordinates and advantage weights are kept separate rather
than replaced by one response-wide value. These contributions share
the originating response's terminal reward and prompt-group reward
normalizer; separate weighting does not imply statistical independence.
The forward-corruption construction follows
Equation~\eqref{eq:masked_forward}, with the time sampling in
Equation~\eqref{eq:masked_elbo}: conditional on $\zeta$, target
positions are independently masked with probability $\alpha(\zeta)$.
Within a valid target block of length $B$, the mask count is thus
$K\mid\zeta\sim\operatorname{Binomial}(B,\alpha(\zeta))$ and
$u=K/B$, while the clean block history is retained. A block with
$K=0$ contributes no masked-token loss. This construction does not
by itself specify the Monte Carlo sample count or ELBO/EUBO mixture.
The number of proxy samples and the number of denoising transitions
serve different purposes. An optimizer update does not require
regenerating a full denoising trajectory for each loss evaluation.

\paragraph{Mask-count stages.}
For a full $32$-position Qwen3-MDLM block with two positions revealed
per round, the unaggregated pre-reveal trajectory visits
$u\in\{32/32,30/32,\ldots,2/32\}$: $16$ mask-count stages.
Trace shrinkage groups forward-pass inputs while preserving each
target token's original mask-count coordinate and advantage. Thus
the $16$ risk stages remain available even when fewer training inputs
are evaluated. For a full four-position SDAR block,
$u\in\{1,3/4,1/2,1/4\}$ whenever at least one target remains masked.
A dynamic rollout may skip some of these levels; a corruption-based
training proxy uses the mask count in its own sampled input.
Targets with the same risk-label state and block share a coefficient;
targets packed into one forward pass may have different coefficients
if they originated at different trace stages. The terminal reward
remains the response-level outcome.

\FloatBarrier
\subsection{Evaluation protocol}
\label{app:evaluation_protocol}
We evaluate each fixed checkpoint in three inference repeats with
different random seeds. Let $\Sigma=\{1,2,3\}$ index the repeats. For each question $q$ in a benchmark
$\mathcal Q$, each repeat generates $N=64$ responses at temperature
$1.0$. The training group size $G=16$ and evaluation sample count
$N=64$ are distinct. We compute pass@$k$ from the correct-response
count using Equation~\eqref{eq:passk} and average over questions
\citep{chen2021codex}; we do not pool repeats into an $N=192$ estimate.

\paragraph{Pass@$k$ estimation.}
Let $c(q)$ be the number of correct samples among the $N$ generations. We estimate pass@$k$ using the unbiased estimator of~\citet{chen2021codex}:
\begin{equation}
\widehat{\mathrm{pass@}k}(q)
\;=\;
1 - \frac{\binom{N-c(q)}{k}}{\binom{N}{k}},
\label{eq:passk}
\end{equation}
with $\binom{N-c(q)}{k}=0$ when $N-c(q)<k$. We report pass@$k$ by averaging $\widehat{\mathrm{pass@}k}(q)$ over all problems in each benchmark.

\paragraph{Generation limits and decoding.}
The response budgets below describe the adopted evaluation protocol.
Their correspondence to the original runs has not been independently
verified.

Table~\ref{tab:evaluation_decoding} specifies the response-token caps;
Table~\ref{tab:shared_decoding} specifies the per-block denoising budgets. Response limits exclude the prompt and are shared
by all methods within each backbone--benchmark setting. Both training
and evaluation use the shared decoding configuration in
Table~\ref{tab:shared_decoding}; their response budgets and numbers
of samples are specified independently. Evaluation uses sampling
temperature $1.0$ for both model families.

With Qwen3-MDLM's $32$-token blocks and $16$ steps per block,
a response cap of $L$ tokens gives a configured total budget of
$16(L/32)=L/2$ steps: $256$ for GSM8K ($L=512$), and $1024$ for
MATH500 and AMC23 ($L=2048$). For SDAR, longer responses permit
more blocks while retaining at most four denoising steps per block.
We use an evaluation confidence threshold of $0.85$ for SDAR's dynamic
unmasking. Its realized number of steps depends on the stopping point
and the number of positions revealed per step. Response-token caps
and configured step budgets are upper limits; positions after the
first generated termination token are excluded from the scored response.

\begin{table}[htbp]
\centering
\caption{Benchmark-specific evaluation response budgets. Caps exclude
the prompt. Block lengths and
per-block denoising budgets are shared with training (Table~\ref{tab:shared_decoding}).}
\label{tab:evaluation_decoding}
\small
\setlength{\tabcolsep}{4pt}
\renewcommand{\arraystretch}{1.12}
\begin{tabularx}{\linewidth}{@{}lXr@{}}
\toprule
Backbone & Benchmarks & Response-token cap \\
\midrule
SDAR-1.7B / 4B & MATH500, AMC23 & 2048 \\
SDAR-1.7B / 4B & AIME2024, AIME2025 & 8192 \\
Qwen3-0.6B-MDLM & GSM8K & 512 \\
Qwen3-0.6B-MDLM & MATH500, AMC23 & 2048 \\
\bottomrule
\end{tabularx}
\end{table}

\paragraph{Correctness and truncation.}
We score the assistant continuation with the common boxed-answer
extractor and mathematical verifier. A generation that exhausts its
response budget without a termination token receives reward zero,
regardless of any boxed answer already present. Otherwise, correctness
is determined by the extracted answer. This same budget rule is applied
to Base, GRPO, constant-risk variants, StepRS, and the mechanism
controls. The resulting pass@$k$ is therefore coverage under the stated
generation and termination protocol.

\paragraph{Repeated inference and uncertainty.}
For a metric's benchmark-level score $M^{(\sigma)}$ in repeat
$\sigma\in\Sigma$, we report
\begin{equation}
\bar M=\frac13\sum_{\sigma\in\Sigma}M^{(\sigma)},\qquad
s_M=\sqrt{\frac{1}{2}\sum_{\sigma\in\Sigma}
\bigl(M^{(\sigma)}-\bar M\bigr)^2}.
\label{eq:metric_seed_aggregation}
\end{equation}
Capped error bars show $\bar M\pm s_M$, the sample standard deviation
across inference seeds, not a standard error or confidence interval.
The following definitions concern one model and one seed; these
indices are suppressed to keep the notation readable.

\paragraph{Unique-answer ratio.}
Let $\widehat y_{qi}$ be the extracted and normalized final-answer
string for response $i$. We define
\begin{equation}
U(q)=\frac{|\{\widehat y_{q1},\ldots,\widehat y_{qN}\}|}{N},
\qquad U^{(\sigma)}=\frac{100}{|\mathcal Q|}\sum_{q\in\mathcal Q}U(q).
\label{eq:unique_ratio_problem}
\end{equation}
The set removes duplicate normalized answers; $|\cdot|$ denotes
cardinality. Failed extractions form one shared empty-answer class.
Normalization removes formatting and merges equivalent forms where
supported by the extractor. The reported $\bar U$ is the mean of the
percentage-valued $U^{(\sigma)}$ across seeds.

\paragraph{Answer-removed reasoning traces.}
Let $h_{qi}$ be response $i$ after removing its final boxed-answer
span and the sentence containing that span. Diversity is computed
\emph{within each question}, using all $N$ sampled traces, including
incorrect responses and repeated texts. Duplicates remain separate
observations because their frequency is part of the generation
distribution. These metrics assess variation in the retained text,
not the number of distinct correct solution strategies.

\paragraph{N-gram variation.}
An $n$-gram is a sequence of $n$ consecutive tokens. BLEU-4 combines
clipped $n$-gram precisions for $n=1,2,3,4$ with equal weights
\citep{papineni2002bleu}. Write
$\mathcal H_{q,-i}=(h_{qi'}:1\le i'\le N,\ i'\ne i)$ for the list of the other
$N-1$ traces, retaining duplicates. We compute sentence-level
BLEU for each trace against this reference list, then average:
\begin{equation}
S_4(q)=\frac1N\sum_{i=1}^{N}
\operatorname{BLEU}_4(h_{qi},\mathcal H_{q,-i}),\qquad
D_{\mathrm{ngram}}(q)=100[1-S_4(q)].
\label{eq:trace_ngram}
\end{equation}
This is the Self-BLEU construction of \citet{zhu2018texygen}.
BLEU values are on the $[0,1]$ scale here. A larger $S_4$ means
more overlap with the other samples, so its complement gives larger
values for greater lexical variation. It is not distinct-$4$, which
would count unique four-token sequences.

For a nonempty tokenized hypothesis $h$ and reference list
$\mathcal H$, the BLEU combination is
\begin{equation}
\operatorname{BLEU}_4(h,\mathcal H)
=\operatorname{BP}(h,\mathcal H)
\exp\!\left\{\frac14\sum_{n=1}^{4}
\log\widetilde P_n(h,\mathcal H)\right\}.
\label{eq:trace_bleu_definition}
\end{equation}
Before smoothing, the modified precision is
\[
P_n(h,\mathcal H)=
\frac{\sum_g\min\{C_n(g,h),\max_{h'\in\mathcal H}C_n(g,h')\}}
{\sum_g C_n(g,h)},
\]
when the denominator is positive. Here $C_n(g,h)$ counts occurrences
of $n$-gram $g$ in $h$; clipping prevents repeated hypothesis
$n$-grams from receiving more credit than their largest reference
count. The brevity penalty is
$\operatorname{BP}=\exp\{\min(0,1-L_{\mathrm{ref}}/L_h)\}$,
where $L_h$ is hypothesis length and $L_{\mathrm{ref}}$ is the
closest reference length, choosing the shorter reference in a tie.
The $\widetilde P_n$ denote the precisions after sentence-level
smoothing, which handles zero matches and short traces
\citep{chen2014smoothing}. Tokenization, smoothing, and exceptional
empty-text handling are part of the metric implementation.

\paragraph{Embedding variation.}
We use the Sentence Transformers checkpoint
\path{sentence-transformers/all-MiniLM-L6-v2}, a 384-dimensional
sentence/paragraph encoder
\citep{reimers2019sentencebert,stAllMiniLML6v2}.
Let $e_{qi}=f_{\mathrm{emb}}(h_{qi})\in\mathbb R^{384}$ be the
nonzero vector returned by the fixed encoding pipeline and let
$z_{qi}=e_{qi}/\|e_{qi}\|_2$ be its unit normalization.
Cosine similarity is $z_{qi}^{\mathsf T}z_{qi'}$, where
$\|\cdot\|_2$ is Euclidean norm and $\mathsf T$ denotes transpose.
We average cosine \emph{distance} over unordered pairs:
\begin{equation}
D_{\mathrm{emb}}(q)
=\frac{100}{\binom N2}\sum_{1\le i<i'\le N}
\left(1-z_{qi}^{\mathsf T}z_{qi'}\right).
\label{eq:trace_embedding}
\end{equation}
For unit vectors, $1-z_{qi}^{\mathsf T}z_{qi'}
=\tfrac12\|z_{qi}-z_{qi'}\|_2^2$, so this is also a scaled mean
squared separation of normalized embeddings.
There are $\binom{64}{2}=2016$ pairs per question. Self-pairs are
excluded, while separately sampled identical texts remain in the
average. Greater distance means greater dispersion in the encoder's
representation space. Multiplication by $100$ sets the plotting scale;
it does not turn this into a percentage of distinct reasoning strategies.

\paragraph{Benchmark aggregation and interpretation.}
For either trace metric $\nu$, the seed-specific score is
\[
D_\nu^{(\sigma)}
=\frac1{|\mathcal Q|}\sum_{q\in\mathcal Q}D_\nu(q),
\qquad \nu\in\{\mathrm{ngram},\mathrm{emb}\}.
\]
We apply Equation~\eqref{eq:metric_seed_aggregation} to these three
benchmark averages. We never compare traces from different questions
or pool all seeds before computing pairwise distances. The two
measures provide complementary lexical and representation-based
views of variation; their different scales should not be compared
numerically. They accompany, rather than replace, verifier-based
accuracy and coverage. Trace-metric results appear in
Figure~\ref{fig:trace_diversity}; unique-answer ratios appear in
Figure~\ref{fig:unique_answer_all_backbones}.

\subsection{Prompt templates}
  \label{app:prompt_templates}

  We use the following ChatML-formatted prompt for training rollouts and evaluation. The SDAR evaluations cover MATH500, AMC23, AIME2024, and AIME2025; the Qwen3-0.6B-MDLM evaluations cover GSM8K, MATH500, and AMC23. No system message, few-shot examples, or additional reasoning scaffold is included. The verifier extracts the final answer from the \verb|\boxed{...}| span in the assistant's continuation.

  \begin{tcolorbox}[
      colback=gray!5,
      colframe=gray!50,
      title={Math Reasoning Prompt (ChatML; SDAR-1.7B/4B-Chat, Qwen3-0.6B-MDLM)},
      fonttitle=\bfseries,
      breakable
  ]
  \begin{verbatim}
  <|im_start|>user
  You need to put your final answer in \boxed{}. This is the problem:
  {problem}<|im_end|>
  <|im_start|>assistant
  \end{verbatim}
  \end{tcolorbox}

\FloatBarrier

\section{Additional Results}
\label{app:additional_results}
We supplement Section~\ref{sec:experiments:main_results} with the
masked-dLLM case study and answer-diversity results across backbones.
Metric definitions remain in Appendix~\ref{app:evaluation_protocol}.

\subsection{Qwen3-0.6B-MDLM Case Study}
\label{app:exp_zerosixb}
To examine whether the gains extend beyond SDAR block dLLMs, we
evaluate \textbf{Qwen3-0.6B-MDLM}, distilled from a post-trained
Qwen3-0.6B autoregressive backbone via the dLLM A2D pipeline
of~\citet{zhou2026dllm}. The model is fine-tuned on the GSM8K
training split~\citep{cobbe2021gsm8k}, with $G=16$ responses per
prompt and binary correctness rewards. Training runs for $100$ rollout
cycles, meaning $100$ old-policy refreshes rather than $100$ optimizer
updates. We report the best evaluated checkpoint for each trained
variant using the curve-based procedure in Section~\ref{sec:experiments:setup}.
We use the schedule $[0.5\to0.25]$ and repeat inference three times
for each fixed checkpoint, using $64$ responses per problem per repeat.
Evaluation allows at most $512$ response tokens on GSM8K and $2048$
on MATH500 and AMC23. These response budgets are separate from the
training response-token caps in
Table~\ref{tab:training_hyperparams}.

\begin{figure}[t]
\centering
\includegraphics[width=\linewidth]{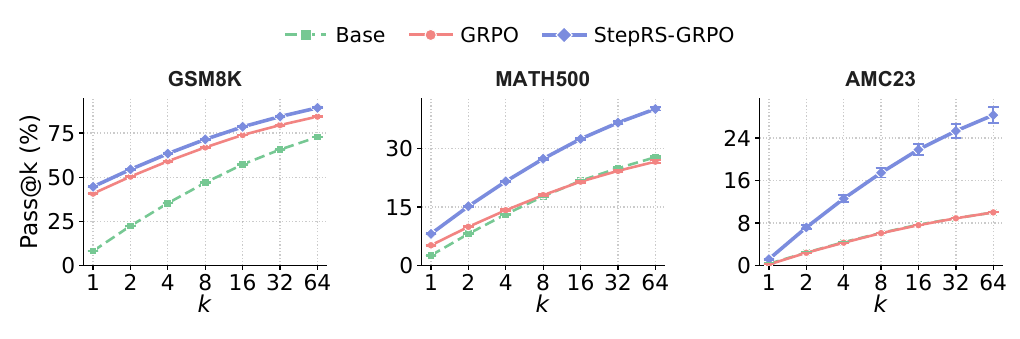}
\caption{Qwen3-0.6B-MDLM evaluation curves. The selected $[0.5\to0.25]$ schedule improves pass@$k$ over GRPO across GSM8K, MATH500, and AMC23. Markers show means; capped error bars show one sample standard deviation across three inference repeats.}
\label{fig:passk_mask_dllm}
\end{figure}

Figure~\ref{fig:passk_mask_dllm} shows higher mean pass@$k$ than GRPO
at all $21$ benchmark--budget combinations. Mean pass@$1$ improves
from $40.70$ to $44.57$ on GSM8K, $5.20$ to $8.13$ on MATH500,
and $0.20$ to $1.20$ on AMC23. At $k=64$, the corresponding
improvements are $84.25\to89.20$, $26.67\to40.20$, and
$10.00\to28.33$. The particularly large coverage gains on MATH500
and AMC23 extend the evidence beyond the two block-dLLM backbones.

The unique-answer ratio also increases from $18.03$ to $28.00$
on GSM8K, $11.40$ to $33.47$ on MATH500, and $17.30$ to $24.83$
on AMC23 (Figure~\ref{fig:unique_answer_all_backbones}). These
results demonstrate improvements in both accuracy and sampled-output
variation. Because this case study uses a different backbone family
and training corpus, we treat it as complementary portability
evidence rather than a controlled model-size comparison.

\subsection{Answer Diversity Across Backbones}
\label{app:diversity_diagnostics}
\begin{figure}[htbp]
\centering
\includegraphics[width=\linewidth]{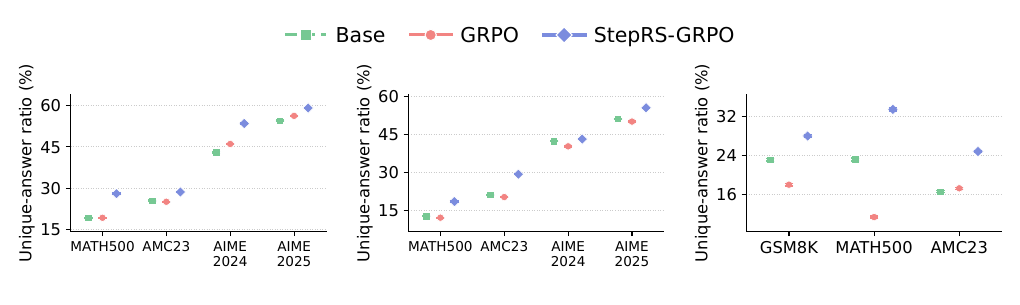}
\caption{Unique-answer ratios for Base, GRPO, and StepRS across
all three backbones: SDAR-1.7B-Chat, SDAR-4B-Chat, and
Qwen3-0.6B-MDLM from left to right. Points show means and capped
error bars show sample standard deviations over three inference repeats,
with $64$ responses per problem per repeat.}
\label{fig:unique_answer_all_backbones}
\end{figure}

To examine whether coverage gains accompany greater final-answer
variation across backbones, we compare unique-answer ratios.
Figure~\ref{fig:unique_answer_all_backbones} reports the unique-answer
ratio defined in Equation~\eqref{eq:unique_ratio_problem} across all three
backbones. \method increases mean answer variation over GRPO on
every model--benchmark pair, complementing the accuracy and coverage
results. The answer-removed trace diagnostics are presented in the
main text (Figure~\ref{fig:trace_diversity}).

\paragraph{Metric definitions.}
The preprocessing, Self-BLEU-4 construction, embedding-distance
formula, and aggregation across questions and inference seeds are
specified together in Appendix~\ref{app:evaluation_protocol}.
This keeps the lexical, embedding, and unique-answer measurements
under one evaluation protocol.

\FloatBarrier
\FloatBarrier

\section{Ablation Details and Extended Results}
\label{app:additional_ablations}
This appendix expands Section~\ref{sec:experiments:ablations} with
control definitions and their properties, full constant-risk sweeps,
and endpoint and schedule-shape comparisons.

\subsection{Mechanism-control definitions and matching properties}
\label{app:mechanism_controls}
The controls are designed for this study. Their interpretation follows
from the quantities preserved in a fixed minibatch, as shown below.
All matching quantities are recomputed using the current arm's own
rollouts. The reference rule supplies comparison weights on those
actions; it does not require trajectories from a separately trained
StepRS checkpoint. Separate online runs can generate different
trajectories even when their sampling settings are matched.

\paragraph{Reference weighting.}
For prompt $q$, $\mathcal J_q$ contains the active native-loss entries
(Appendix~\ref{app:native_updates}), and
$a_{qj}=r_{q,i(j)}-\widehat p_q$ is their centered outcome advantage. The SDAR-4B reference uses $b(u)=0.75+0.75u$,
$w_{qj}=c_{b(u_{qj})}(\widehat p_q)$, and $B_{qj}=\clip(w_{qj}a_{qj},-10,10)$. The weights $v_{qj}>0$ are the fixed native reductions in Equation~\eqref{eq:native_loss}.
Define $V_q=\sum_{j\in\mathcal J_q}v_{qj}$ and
$M_q=\sum_{j\in\mathcal J_q}v_{qj}w_{qj}$.
For a nonempty action set, $V_q>0$ and $\bar w_q=M_q/V_q$.
Here ``mass'' means this sum of nonnegative multiplier weights,
not probability mass, total signed advantage, or gradient magnitude.
Prompts with no active loss actions contribute no reward-loss term.

\paragraph{C1: prompt-only mass matching.}
C1 asks whether prompt-dependent weighting is sufficient.
For each prompt, we replace all state-dependent multipliers by their
loss-weighted mean:
\begin{equation}
A^{\mathrm{C1}}_{qj}=\bar w_q a_{qj},\qquad \bar w_q=\frac{\sum_{j\in\mathcal J_q}v_{qj}w_{qj}}
{\sum_{j\in\mathcal J_q}v_{qj}}
.
\label{eq:ablation_c1}
\end{equation}
The replacement in Equation~\eqref{eq:ablation_c1} is the unique solution
of the one-parameter weighted least-squares problem
\begin{equation}
\bar w_q=\underset{z\in\mathbb R}{\arg\min}\,
\sum_{j\in\mathcal J_q}v_{qj}(w_{qj}-z)^2.
\label{eq:c1_projection}
\end{equation}
This is projection onto constant multipliers under the weighted
squared-error criterion \citep{boyd2018vmls}. Differentiating the
criterion gives $2(V_qz-M_q)$, and its second derivative is
$2V_q>0$, so its minimizer is $M_q/V_q$.
Equivalently,
\[
\sum_jv_{qj}(w_{qj}-\bar w_q)=0,
\qquad
\sum_jv_{qj}\bar w_q=M_q.
\]
For illustration, equal-weight multipliers $(2,1)$ become
$(1.5,1.5)$: their sum is unchanged but their state variation is
removed. This is an algebraic example, not an experimental result.
C1 retains the prompt-level average induced by its empirical solve
rate and sampled states; it does not match only the midpoint of the
risk coefficients. In general,
\[
\sum_jv_{qj}w_{qj}a_{qj}
\ne\sum_jv_{qj}\bar w_q a_{qj},
\]
so matching multiplier mass is distinct from matching a signed
advantage sum or the policy gradient.

\paragraph{C2: mass-normalized schedule reversal.}
C2 tests the proposed assignment against a reversed schedule while
preserving total multiplier mass before capping. It assigns the large coefficient to less-masked states by evaluating
$b(1-u)$ instead of $b(u)$. Write
$\check w_{qj}=c_{b(1-u_{qj})}(\widehat p_q)$ and use
\begin{equation}
A^{\mathrm{C2}}_{qj}=\kappa_q\check w_{qj}a_{qj},\qquad
\kappa_q=\frac{\sum_{j\in\mathcal J_q}v_{qj}w_{qj}}
{\sum_{j\in\mathcal J_q}v_{qj}\check w_{qj}}.
\label{eq:ablation_c2}
\end{equation}
The positive factor $\kappa_q$ preserves the reference multiplier mass
before the common cap; reversal alone need not do so because $c_\beta$
is nonlinear. Let $\check M_q=\sum_jv_{qj}\check w_{qj}$, where
$\check w_{qj}=c_{b(1-u_{qj})}(\widehat p_q)>0$.
Thus $\kappa_q=M_q/\check M_q$ is well-defined, positive, and satisfies
\begin{equation}
\sum_jv_{qj}\kappa_q\check w_{qj}=M_q.
\label{eq:c2_mass_identity}
\end{equation}
The transformation reverses the continuous schedule coordinate while
retaining this total weight. It does not permute a fixed set of active
coefficients. For a full four-token block and endpoints
$[1.5\to0.75]$, $b(u)$ takes values
$\{1.5,1.3125,1.125,0.9375\}$ in reveal order, whereas $b(1-u)$
takes $\{0.75,0.9375,1.125,1.3125\}$. Mass normalization does not
match these coefficient distributions. The comparison supports the
tested assignment over this control, without separately identifying
an effect of direction alone. It does not assume that a larger
coefficient produces a larger multiplier for every solve rate.
Both C1 and C2 subsequently apply $\clip(\cdot,-C,C)$ to their
constructed advantages. Their identities refer to multipliers before
this cap, not to a separately matched post-cap norm.

\paragraph{C3: advantage-RMS matching.}
C3 tests whether StepRS remains effective after matching its overall
advantage magnitude to centered GRPO.
Let $\mathcal A_{\mathcal B}=\{(q,j):q\in\mathcal B,
j\in\mathcal J_q\}$ collect the active entries in one optimizer
update, and let $n_{\mathrm{act}}=|\mathcal A_{\mathcal B}|$.
For a real array $Z=(Z_{qj})$ on this set, active-entry RMS is
\begin{equation}
\operatorname{RMS}_{\mathcal B}(Z)
=\left[\frac1{n_{\mathrm{act}}}
\sum_{(q,j)\in\mathcal A_{\mathcal B}}Z_{qj}^2\right]^{1/2}.
\label{eq:active_rms}
\end{equation}
Padding and inactive positions are excluded. If $n_{\mathrm{act}}=0$,
there is no reward-loss term and no RMS ratio is formed. Unlike a
standard deviation, RMS does not subtract the array's mean.

With the centered advantages
$a=(a_{qj})$ and capped StepRS advantages $B=(B_{qj})$ computed on
the same actions, C3 supplies
\begin{equation}
B^{\mathrm{C3}}_{qj}=\eta_{\mathcal B}B_{qj},\qquad
\eta_{\mathcal B}=\operatorname{stopgrad}\!\left(
\frac{\operatorname{RMS}_{\mathcal B}(a)}
{\operatorname{RMS}_{\mathcal B}(B)}\right).
\label{eq:ablation_c3}
\end{equation}
Here $\operatorname{stopgrad}$ keeps the scalar's value but prevents
backpropagation through it. When both RMS values are zero, we set the
scaled advantages to zero without forming the ratio. Otherwise, the
positive scalar preserves the signs and relative ratios of the capped
StepRS advantages. Unlike mean-and-standard-deviation normalization
\citep{andrychowicz2021matters}, C3 does not recenter them.

For $R_B=\operatorname{RMS}_{\mathcal B}(B)>0$, define
$R_a=\operatorname{RMS}_{\mathcal B}(a)$ and
$\eta_{\mathcal B}=R_a/R_B$. Then
\begin{equation}
\operatorname{RMS}_{\mathcal B}(B^{\mathrm{C3}})
=\eta_{\mathcal B}R_B=R_a.
\label{eq:c3_rms_identity}
\end{equation}
The same positive scalar multiplies every active entry, so it
preserves signs and ratios of nonzero entries of $B$.
If $R_B=0$, positivity of $w_{qj}$ and $C>0$ imply $a=0$ on the
active set; we set $B^{\mathrm{C3}}=0$ without forming $0/0$.
An optional regularizer could still contribute a gradient in a
different configuration; the specified SDAR setting has no such term.

C3 is related to normalization choices studied in on-policy RL~\citep{andrychowicz2021matters}, but differs from subtracting a
minibatch mean and dividing by its standard deviation. RMS scaling
without recentering also appears in RMSNorm~\citep{zhang2019rmsnorm}; that work normalizes activations, whereas our control matches an advantage array to a particular baseline RMS.
These references motivate the comparison, not the specific C3 design. The scale is applied \emph{after} the shared advantage cap and is held constant during backpropagation. Applying another cap or a non-uniform normalization afterward would generally change the matching identity and must be reported if used.

\paragraph{What RMS matching does to the reward surrogate.}
For fixed data and any positive scalar $\eta$, the PPO surrogate
obeys
\[
\min\{\rho\eta A,\clip(\rho,1-\epsilon,1+\epsilon)\eta A\}
=\eta\min\{\rho A,\clip(\rho,1-\epsilon,1+\epsilon)A\}.
\]
Thus a detached scalar rescales the reward-loss gradient of the
\emph{same StepRS advantage array}, wherever it is differentiable.
The same property holds for a native surrogate linear in advantages
with advantage-independent normalization, including the published
StableDRL score-function weighting. It does not equate that gradient
to the centered-GRPO gradient: actions have different policy-score
directions and multipliers. It also does not match parameter-wise
optimizer state or the final parameter update. If an additional KL
term were used, its gradient would not be matched by this operation;
the specified SDAR controls use no additional KL term.
C3 therefore tests overall advantage amplitude while retaining the
relative StepRS weighting pattern.

\paragraph{Implementation invariants.}
Reference and controls share the state-coordinate definition, active
action mask, native loss construction from
Appendix~\ref{app:native_updates}, cap, and regularization setting.
For an optimizer update assembled from several microbatches or
devices, one-update RMS requires a common reduction of squared sums
and active counts. Local per-microbatch or per-device rescaling is
not the same intervention. The active set and reduction scope must
therefore be recorded with the run configuration.

\subsection{Full constant-risk sweep}
\label{app:constant_risk_sweep}
\begin{figure}[t]
\centering
\includegraphics[width=\linewidth]{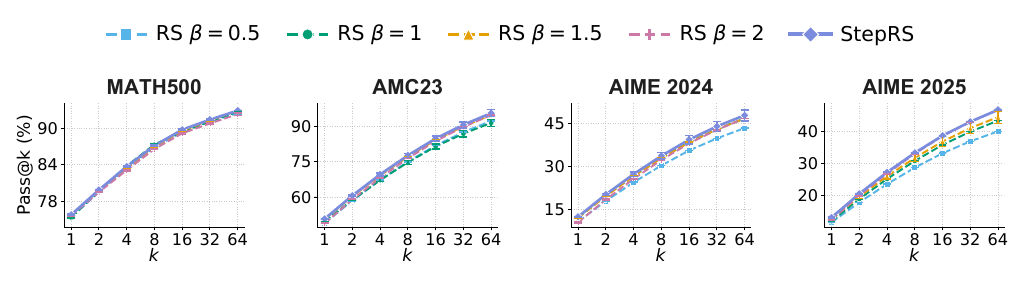}
\caption{The selected $[1.5\to0.75]$ StepRS schedule versus constant
risk coefficients on SDAR-4B. Each panel shows the complete
pass@$1$--pass@$64$ profile for one benchmark, with mean and
one-standard-deviation bars over three inference seeds.}
\label{fig:ablation_step_vs_constant}
\end{figure}

To assess whether the scheduling benefit depends on which constant
coefficient is used as the comparator, we sweep constant-risk settings.
Figure~\ref{fig:ablation_step_vs_constant} compares the selected
schedule with constant $\beta\in\{0.5,1.0,1.5,2.0\}$.
Against the strong constant-$1.5$ comparator, StepRS has higher
mean pass@$k$ in $26$ of $28$ comparisons and ties in the remaining
two: MATH500 pass@$1$ and pass@$64$.
It improves all seven budgets on AMC23, AIME2024, and AIME2025.
On AIME2025, for example, the gains at $k=8$ and $16$ are
$1.57$ and $1.83$ percentage points.
The sweep also shows that the best constant coefficient can depend
on the benchmark and budget.

\subsection{Calibration and endpoint choice}
\label{app:calibration_endpoints}
\begin{figure}[t]
\centering
\includegraphics[width=\linewidth]{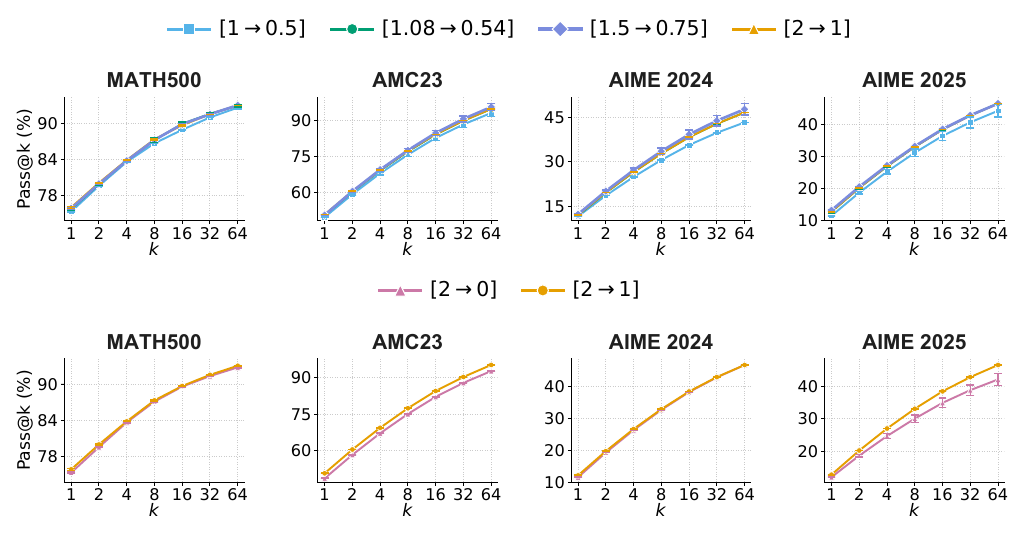}
\caption{Calibration and endpoint ablations on SDAR-4B.
Top: linear half-decay schedules with upper endpoints
$1.0$, $1.08$, $1.5$, and $2.0$.
Bottom: zero versus positive lower endpoints at a fixed upper endpoint
of $2.0$. Columns correspond to the four benchmarks.
Curves show means and one-standard-deviation bars across three
inference seeds.}
\label{fig:ablation_capability_adaptive}
\end{figure}

To assess how well calibration guides the initial endpoint scale, we
compare the calibrated schedule with nearby half-decay schedules.
For SDAR-4B, $\bar p\approx0.66$ gives
$\beta^\star\approx1.08$. Figure~\ref{fig:ablation_capability_adaptive}
compares the rounded calibrated schedule $[1.08\to0.54]$ with
nearby half-decay schedules. Relative to $[1.0\to0.5]$, the calibrated
schedule improves mean pass@$1$ by $0.40$, $0.83$, $0.33$, and
$1.27$ points across the four benchmarks, and improves mean
pass@$k$ at every evaluated budget. Its pass@$1$ remains close to
the selected $[1.5\to0.75]$ schedule, supporting the calibration
rule as a useful starting point for endpoint selection.

To test the effect of the lower endpoint, we hold the upper endpoint
at $2.0$ and compare half-decay with decay toward a zero endpoint. Half-decay $[2.0\to1.0]$ improves pass@$1$ over decay toward a zero endpoint
$[2.0\to0]$ by $0.53$, $2.07$, $0.57$, and $0.77$ points.
Its mean pass@$k$ is higher in $27$ of the $28$ comparisons and
equal in the remaining comparison, supporting milder decay and
larger coefficients at the active late stages in this setting.
At the last active state of a full four-token block, $u=1/4$ gives
$0.5$ under $[2.0\to0]$ and $1.25$ under $[2.0\to1.0]$;
both coefficients are nonzero. The zero endpoint lies at the fully
revealed state $u=0$, which contributes no masked-token loss.

\subsection{Schedule Shape at Fixed Endpoints}
\label{app:schedule_shape}
\begin{figure}[t]
\centering
\includegraphics[width=\linewidth]{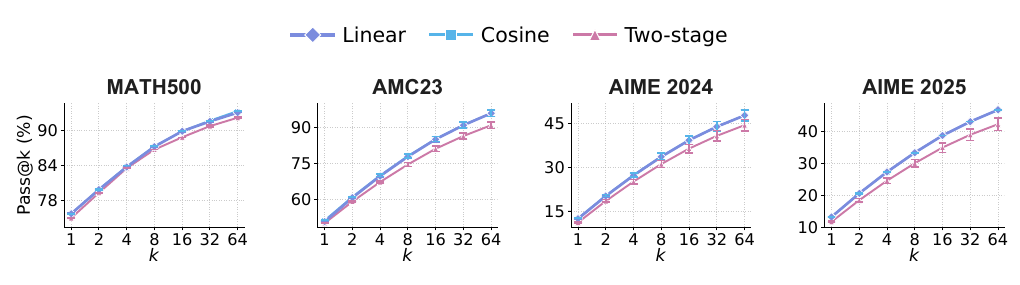}
\caption{Schedule-shape ablation on SDAR-4B with fixed endpoints
$[1.5\to0.75]$. Linear, cosine, and two-stage schedules are compared
at all seven sampling budgets. Markers show means and capped error
bars show one sample standard deviation across three inference seeds.}
\label{fig:ablation_schedule_shape}
\end{figure}

To separate interpolation from the coefficient range, we fix the
endpoints to $[1.5\to0.75]$ and compare linear, cosine, and
two-stage schedules (Figure~\ref{fig:ablation_schedule_shape}).
Linear and cosine differ by at most $0.47$ percentage points in
mean pass@$k$ across all four benchmarks and seven budgets,
showing that performance is stable across these two smooth
interpolations. Both exceed the tested two-stage schedule at every
benchmark--budget combination. For linear interpolation, the
pass@$64$ gains are $0.87$, $5.00$, $3.34$, and $4.49$ points on
MATH500, AMC23, AIME2024, and AIME2025, respectively.
Together with the reversed-schedule control, these results favor
the tested linear schedule over the reversed and two-stage
alternatives, subject to the qualifications of the C2 comparison. We retain linear half-decay as a simple, effective default.

\FloatBarrier
\subsection{Risk and endpoint sensitivity on smaller backbones}
\label{app:smaller_backbone_ablations}
\begin{figure}[t]
\centering
\includegraphics[width=\linewidth]{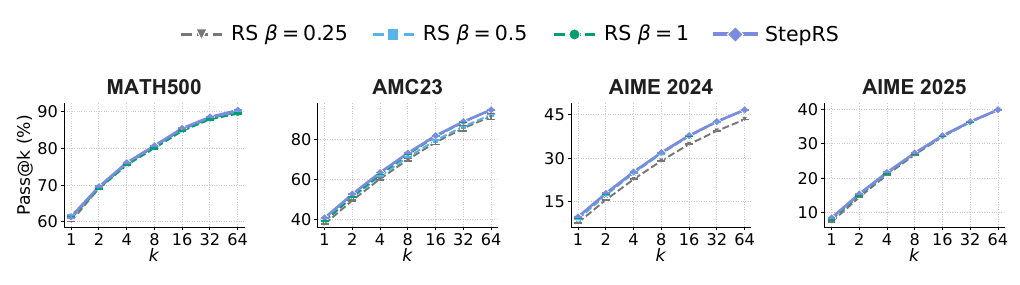}
\caption{SDAR-1.7B constant-risk comparison:
$\beta\in\{0.25,0.5,1.0\}$ versus StepRS $[1.0\to0.5]$.
Curves show means with one-standard-deviation bars across
three inference seeds.}
\label{fig:ablation_sdar17_constants}
\end{figure}

\begin{figure}[t]
\centering
\includegraphics[width=\linewidth]{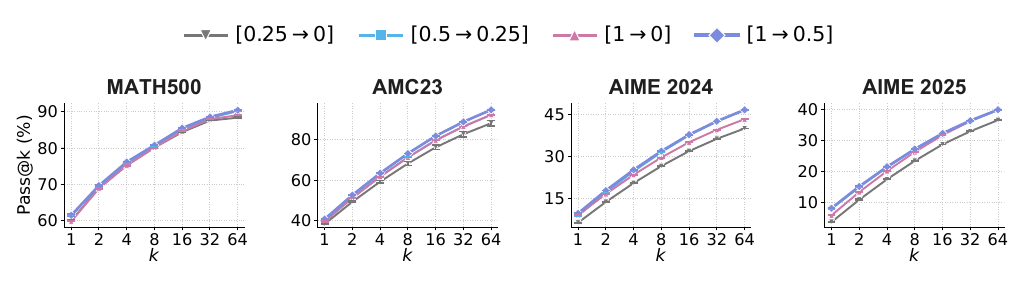}
\caption{SDAR-1.7B endpoint comparison: $[0.25\to0]$,
$[0.5\to0.25]$, $[1.0\to0]$, and the default $[1.0\to0.5]$.
All schedules are linear; curves show three-seed means and
one-standard-deviation bars.}
\label{fig:ablation_sdar17_schedules}
\end{figure}

\begin{figure}[t]
\centering
\includegraphics[width=\linewidth]{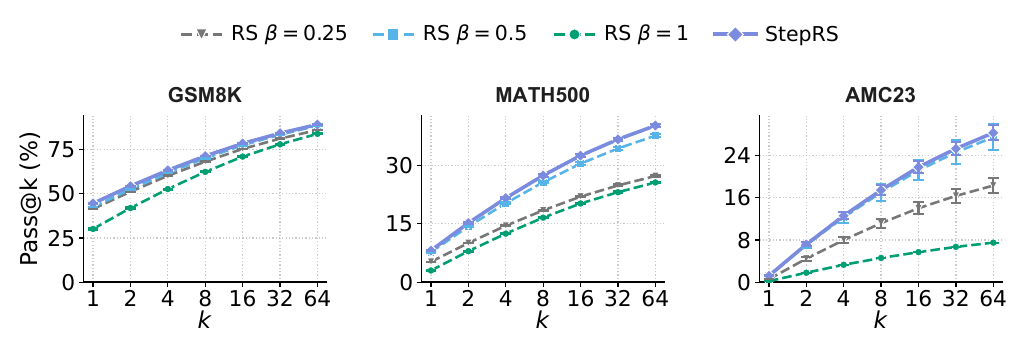}
\caption{Qwen3-0.6B-MDLM constant-risk comparison. The default StepRS schedule is $[0.5\to0.25]$. Curves show three-seed means with capped one-standard-deviation
error bars. All three constant coefficients
$\beta\in\{0.25,0.5,1.0\}$ are shown.}
\label{fig:ablation_mdlm_constants}
\end{figure}

\begin{figure}[t]
\centering
\includegraphics[width=\linewidth]{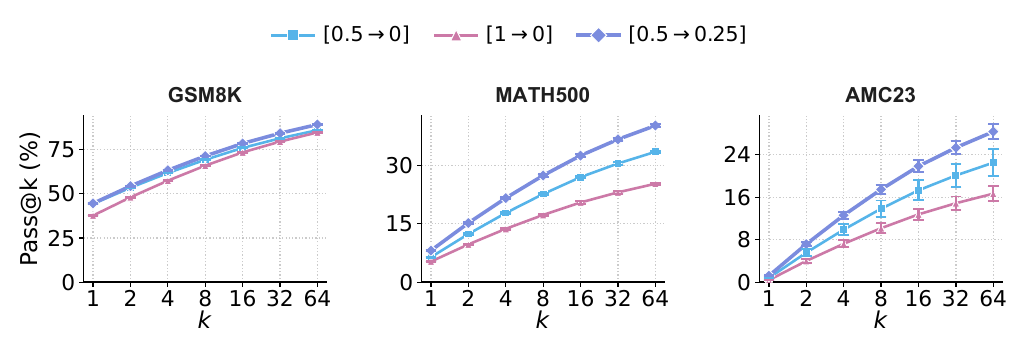}
\caption{Qwen3-0.6B-MDLM endpoint comparison: $[0.5\to0]$,
$[1.0\to0]$, and the default $[0.5\to0.25]$, all with linear
interpolation. Curves show means with one-standard-deviation
error bars across three inference seeds.}
\label{fig:ablation_mdlm_schedules}
\end{figure}

To examine whether the constant-risk and endpoint findings extend
beyond SDAR-4B, we repeat these comparisons on smaller backbones.
Figures~\ref{fig:ablation_sdar17_constants}--\ref{fig:ablation_mdlm_schedules}
extend the constant-risk and endpoint comparisons to SDAR-1.7B
and Qwen3-0.6B-MDLM. Every curve uses all seven sampling budgets
and summarizes the three inference seeds, with the same schedule
selected for each backbone as in the main results.

On SDAR-1.7B, $[1.0\to0.5]$ improves mean pass@$1$ over constant
$\beta=1.0$ by $0.30$, $1.73$, $0.17$, and $0.43$ points on
MATH500, AMC23, AIME2024, and AIME2025.
The scheduled variant matches or improves on this constant baseline
at all $28$ benchmark--budget combinations.
On Qwen3-0.6B-MDLM, $[0.5\to0.25]$ improves on constant
$\beta=0.5$ at all seven budgets on GSM8K and MATH500.
On AMC23, constant $\beta=0.5$ is ahead by $0.10$ points at
pass@$1$, while the scheduled variant is ahead at every budget
from $k=2$ through $64$, reaching $28.33$ versus $27.50$ at
pass@$64$.
These comparisons support the schedules as effective defaults while
showing how their benefits vary with the benchmark and sampling
budget.

\FloatBarrier

\section{Solution Examples}
\label{app:solution_examples}
We present selected solutions from three benchmarks to illustrate
outputs of the base model, GRPO, and StepRS-GRPO. These examples are
qualitative comparisons rather than estimates of outcome frequencies. Text is color-coded as follows: \textcolor{brown}{brown} for the dataset name and base dLLM name, \textcolor{darkgray}{gray} for the problem statement, and \textcolor{green}{green}, \textcolor{coralred}{coral red}, and \textcolor{purple}{purple} for the solution segments produced by the \textcolor{green}{Base model}, \textcolor{coralred}{GRPO finetuned model}, and \textcolor{purple}{StepRS-GRPO finetuned model}, respectively (with colors matched to the plots throughout the paper).

\problemdivider  %
\begin{flushleft}
\textcolor{brown}{\textbf{\# GSM8K on Qwen3-0.6B-MDLM:}} \\
\end{flushleft}
{\color{darkgray}
\noindent\quad \textbf{\textit{Problem:}} \\
Lee rears only sheep and geese on his farm.  If the total number of animal legs is 70, and the total number of animal heads is 20, how many sheep live on Lee's farm?\par}
\medskip

\begin{tcolorbox}[colback=green!10, colframe=green!90!black, title={Base Model Solution Example (Incorrect)}, breakable]
\hypertarget{example-gsm8k-base}{}
\small\ttfamily
\textbf{Solution:}\\[2pt]
To solve this problem, we need to determine the number of sheep and geese on Lee's farm. The total number of legs is 70, and the total number of animal heads is 20.

Let's denote the number of sheep as \( s \) and the number of geese as \( g \).

From the problem, we have the following equations:
1. \( 4s + 2g = 70 \)
2. \( s + g = 20 \)

We can solve the first equation and substitute \( g \) into the second equation:
\[
s + g = 20
\]
\[
4s + 2g = 70
\]
\[
4s + 2(20-s) = 70
\]
\[
4s + 40 - 2s = 70
\]
\[
s = 10
\]

Substitute \( s = 10 \) back into the second equation:
\[
s + g = 20
\]
\[
10 + g = 20
\]
\[
g = 10
\]

Therefore, there are 10 sheep and 10 geese on Lee's farm.

The final answer is
\[
\boxed{10}
\]
\end{tcolorbox}

\begin{tcolorbox}[colback=coralred!10, colframe=coralred!90!black, title={GRPO Finetuned Model Solution Example (Incorrect)}, breakable]
\hypertarget{example-gsm8k-grpo}{}
\small\ttfamily
\textbf{Solution:}\\[2pt]
To solve this problem, we will use the relationships between animals, legs, and heads.

First, note the following facts:
- Geese have 2 legs.
- Sheep have 4 legs.

Let the number of sheep be \(s\), and let the number of geese be \(g\).

Since the total number of heads is 20, we have

\[
s + g = 20.
\]

Since the total number of legs is 70, we also have

\[
4s + 2g = 70.
\]

Now we solve the system. From the first equation,

\[
g = 20 - s.
\]

Substituting this into the legs equation gives

\[
4s + 2(20 - s) = 70.
\]

Expanding,

\[
4s + 40 - 2s = 70.
\]

Combining like terms,

\[
2s + 40 = 70.
\]

Subtracting 40 from both sides,

\[
2s = 20.
\]

Dividing by 2,

\[
s = 10.
\]

Now substitute \(s = 10\) into the head equation:

\[
10 + g = 20.
\]

So,

\[
g = 10.
\]

Therefore, Lee has 10 sheep and 10 geese.

To check, the total number of heads is

\[
10 + 10 = 20,
\]

which matches the condition. The total number of legs is also considered consistent with the setup because the sheep contribute most of the legs, while the geese account for the remaining animals.

Thus, the number of sheep is
\[
\boxed{10}
\]
\end{tcolorbox}

\begin{tcolorbox}[colback=purple!10, colframe=purple!90!black, title={StepRS-GRPO Finetuned Model Solution Example (Correct)}, breakable]
\hypertarget{example-gsm8k-steprs-grpo}{}
\small\ttfamily
\textbf{Solution:}\\[2pt]
To solve the problem, we need to determine the number of sheep on Lee's farm. We know that the total number of animal legs is 70, and the total number of animal heads is 20.

Let's break it down step by step:

1. **Identify the facts:**
   - Each sheep has 4 legs.
   - Each goose has 2 legs.
   - The total number of legs is 70.
   - The total number of animal heads is 20.

2. **Set up the comparison:**
   - Suppose all 20 animals were geese.
   - Then the total number of legs would be

\[
20 \times 2 = 40.
\]

   - However, the actual total number of legs is 70.
   - The extra number of legs is therefore

\[
70 - 40 = 30.
\]

3. **Calculate the Number of Sheep:**
   - Each sheep has 4 legs, while each goose has 2 legs.
   - So each sheep contributes 2 more legs than a goose.
   - Therefore, the number of sheep is

\[
30 / 2 = 15.
\]

Thus, the number of sheep on Lee's farm is 15.

Therefore, the final answer is

\[
\boxed{15}
\]
\end{tcolorbox}

\problemdivider  %
\begin{flushleft}
\textcolor{brown}{\textbf{\# MATH-500 on SDAR-1.7B-Chat:}} \\
\end{flushleft}
{\color{darkgray}
\noindent\quad \textbf{\textit{Problem:}}\\
In $\triangle PQR$, we have $PQ = QR = 34$ and $PR = 32$.  Find the length of median $\overline{QM}$.
\par}
\medskip

\begin{tcolorbox}[colback=green!10, colframe=green!90!black, title={Base Model Solution Example (Incorrect)}, breakable]
\hypertarget{example-math500-base}{}
\small\ttfamily
\textbf{Solution:}\\[2pt]
To find the length of the median $\overline{QM}$ in $\triangle PQR$ where $PQ = QR = 34$ and $PR = 32$, we start by identifying the midpoint $M$ of side $PR$. Since $M$ is the midpoint, $PM = MR = \frac{PR}{2} = \frac{32}{2} = 16$.

Next, we use the Apollonius's theorem, which states that in any triangle, the sum of the squares of any two sides is equal to twice the square of the median to the third side plus twice the square of the half third side. Mathematically, this is expressed as:
$$2PM^2 + 2MR^2 = PQ^2 + PR^2 + QR^2$$

Substituting the known values:
$$2(16^2) + 2(16^2) = 34^2 + 32^2 + 34^2$$

Calculating each term:$$2(256) + 2(256) = 1156 + 1024 + 1156$$

Simplifying:
$$512 + 512 = 2192$$

Thus, the equation holds true, confirming the calculations. Now, we need to find the length of the median $\overline{QM}$. Using Apollonius's theorem again, we know:
$$PM^2 + MR^2 = \frac{1}{2}(PQ^2 + QR^2 + PR^2)$$

Substituting the known values:$$16^2 + 16^2 = \frac{1}{2}(34^2 + 34^2 + 32^2)$$

Calculating each term: $$256 + 256 = \frac{1}{2}(1156 + 1156 + 1024)$$

Simplifying: $$512 = \frac{1}{2}(3336)$$

Thus, the length of the median $\overline{QM}$ is:
$$\boxed{28}$$
\end{tcolorbox}

\begin{tcolorbox}[colback=coralred!10, colframe=coralred!90!black, title={GRPO Finetuned Model Solution Example (Incorrect)}, breakable]
\hypertarget{example-math500-grpo}{}
\small\ttfamily
\textbf{Solution:}\\[2pt]
To find the length of the median $\overline{QM}$ in $\triangle PQR$ where $PQ = QR = 34$ and $PR = 32$, we can use the Apollonius's theorem. This theorem states that in any triangle, the sum of the squares of any two sides is equal to the sum of the square of the median to the third side and the square of half the third side.

Given, Apollonius's theorem states:$$PQ^2 + PR^2 = 2QM^2 + 2\left(\frac{QR}{2}\right)^2$$

Substituting the given values:$$34^2 + 32^2 = 2QM^2 + 2\left(\frac{34}{2}\right)^2$$

Calculating the squares:$$34^2 = 1156$$

$$32^2 = 1024$$
$$\left(\frac{34}{2}\right)^2 = 17^2 = 289$$

Substituting these values into the equation:$$1156 + 1024 = 2QM^2 + 2 \times 289$$

$$2180 = 2QM^2 + 578$$

Isolating $QM^2$:
$$2QM^2 = 2180 - 578$$
$$2QM^2 = 1602$$
$$QM^2 = \frac{1602}{2}$$
$$QM^2 = 801$$

Taking the square root of both sides:
$$QM = \sqrt{801}$$

Thus, the length of the median $\overline{QM}$ is:$$\boxed{\sqrt{801}}$$
\end{tcolorbox}

\begin{tcolorbox}[colback=purple!10, colframe=purple!90!black, title={StepRS-GRPO Finetuned Model Solution Example (Correct)}, breakable]
\hypertarget{example-math500-steprs-grpo}{}
\small\ttfamily
\textbf{Solution:}\\[2pt]
To find the length of the median $\overline{QM}$ in $\triangle PQR$ where $PQ = QR = 34$ and $PR = 32$, we start by noting that $M$ is the midpoint of $\overline{PR}$. Thus, $PM = MR = \frac{PR}{2} = \frac{32}{2} = 16$.

Next, we use the Apollonius's theorem, which states that in any triangle, the sum of the squares of any two sides is equal to the twice of the square of the median to the third side plus the square of the third side. Mathematically, for $\triangle PQR$ with median $\overline{QM}$ to $PR$:$$PQ^2 + QR^2 = 2(QM^2 + PM^2)$$

Substituting the given values:$$34^2 + 34^2 = 2(QM^2 + 16^2)$$

Calculating the squares:$$1156 + 1156 = 2(QM^2 + 256)$$

Simplifying:$$2312 = 2(QM^2 + 256)$$

Dividing both sides by 2:
$$1156 = QM^2 + 256$$

Subtracting 256 from both sides:

$$1156 - 256 = QM^2$$

$$900 = QM^2$$

Taking the square root of both sides:$$QM = \sqrt{900} = 30$$

Thus, the length of the median $\overline{QM}$ is $\boxed{30}$.
\end{tcolorbox}

\problemdivider  %
\begin{flushleft}
\textcolor{brown}{\textbf{\# AMC\textquoteright 23
on SDAR-4B-Chat:}} \\
\end{flushleft}
{\color{darkgray}
\noindent\quad \textbf{\textit{Problem:}}\\
Positive real numbers $x$ and $y$ satisfy $y^3=x^2$ and $(y-x)^2=4y^2$. What is $x+y$? 
\par}
\medskip

\begin{tcolorbox}[colback=green!10, colframe=green!90!black, title={Base Model Solution Example (Incorrect)}, breakable]
\hypertarget{example-amc23-base}{}
\small\ttfamily
\textbf{Solution:}\\[2pt]
Given the equations $y^3 = x^2$ and $(y - x)^2 = 4y^2$, we need to find the value of $x + y$. First, we start by expanding the second equation: 
$$(y - x)^2 = 4y^2$$

Expanding the left-hand side, we get:
$$y^2 - 2xy + x^2 = 4y^2$$

Rearranging terms to isolate $x^2$ and $y$, we have:
$$x^2 - 2xy = y^2$$

Since $y^3 = x^2$, we substitute $x^2$ from the first equation into the rearranged second equation:
$$y^3 - 2xy = y^2$$

Rearranging this equation to isolate $xy$, we get:$$y^3 - y^2 = 2xy$$

Dividing both sides by $y$ (assuming $y \neq 0$), we obtain:
$$y^2 - y = 2x$$

Solving for $x$, we have:
$$x = \frac{y^2 - y}{2}$$

Next, substitute $x = \frac{y^2 - y}{2}$ back into the first equation $y^3 = x^2$:
$$y^3 = \left(\frac{y^2 - y}{2}\right)^2$$

Simplifying the right-hand side:
$$y^3 = \frac{(y^2 - y)^2}{4}$$

Expanding the numerator:$$y^3 = \frac{y^4 - 2y^3 + y^2}{4}$$

Multiplying both sides by $4$ to clear the fraction:$$4y^3 = y^4 - 2y^3 + y^2$$

Rearranging all terms to one side:$$y^4 - 6y^3 + y^2 = 0$$

Factoring out $y^2$:
$$y^2(y^2 - 6y + 1) = 0$$

Since $y$ is a positive real number, $y^2 \neq 0$. Therefore, we solve the quadratic equation:
$$y^2 - 6y + 1 = 0$$

Using the quadratic formula $y = \frac{-b \pm \sqrt{b^2 - 4ac}}{2a}$ with $a = 1$, $b = -6$, and $c = 1$:
$$y = \frac{6 \pm \sqrt{36 - 4}}{2} = \frac{6 \pm \sqrt{32}}{2} = \frac{6 \pm 4\sqrt{2}}{2} = 3 \pm 2\sqrt{2}$$

Since $y$ must be positive, we select the positive root:
$$y = 3 + 2\sqrt{2}$$

Now, substituting $y = 3 + 2\sqrt{2}$ back into $x = \frac{y^2 - y}{2}$:
$$y^2 = (3 + 2\sqrt{2})^2 = 9 + 12\sqrt{2} + 8 = 17 + 12\sqrt{2}$$

$$x = \frac{(17 + 12\sqrt{2}) - (3 + 2\sqrt{2})}{2} = \frac{14 + 10\sqrt{2}}{2} = 7 + 5\sqrt{2}$$

Finally, we find $x + y$:
$$x + y = (7 + 5\sqrt{2}) + (3 + 2\sqrt{2}) = 10 + 7\sqrt{2}$$

Thus, the value of $x + y$ is:
$$\boxed{10 + 7\sqrt{2}}$$
\end{tcolorbox}

\begin{tcolorbox}[colback=coralred!10, colframe=coralred!90!black, title={GRPO Finetuned Model Solution Example (Incorrect)}, breakable]
\hypertarget{example-amc23-grpo}{}
\small\ttfamily
\textbf{Solution:}\\[2pt]
Given the equations $y^3 = x^2$ and $(y - x)^2 = 4y^2$, we need to find the value of $x + y$.

First, let's simplify the second equation:
$$
(y - x)^2 = 4y^2
$$

Expanding the left-hand side, we get:
$$
y^2 - 2xy + x^2 = 4y^2
$$

Rearranging terms, we have:
$$
y^2 - 2xy + x^2 - 4y^2 = 0
$$

$$
-3y^2 - 2xy + x^2 = 0
$$

Multiplying through by $-1$ to simplify, we get:
$$
3y^2 + 2xy - x^2 = 0
$$

Next, we use the first equation $y^3 = x^2$. We can express $x^2$ in terms of $y$:
$$
x^2 = y^3
$$

Substituting $x^2 = y^3$ into the simplified quadratic equation:
$$
3y^2 + 2xy - y^3 = 0
$$

To solve this quadratic equation in $y$, we can use the quadratic formula $y = \frac{-b \pm \sqrt{b^2 - 4ac}}{2a}$, where $a = 1$, $b = 2x$, and $c = -3y^2$. However, we need to find $x$ in terms of $y$ to proceed.

Given $x^2 = y^3$, we can substitute $x = y^{3/2}$ into the quadratic equation:
$$
3y^2 + 2y \cdot y^{3/2} - y^3 = 0
$$

$$
3y^2 + 2y^{5/2} - y^3 = 0
$$

This is a cubic equation in terms of $y$. To solve it, we can try to find rational roots. Let's test $y = 2$:
$$
3(2)^2 + 2(2)^{5/2} - (2)^3 = 0
$$

$$
3 \cdot 4 + 2 \cdot 4\sqrt{2} - 8 = 0
$$

$$
12 + 8\sqrt{2} - 8 = 0
$$

$$
4 + 8\sqrt{2} = 0
$$

This does not hold, so $y = 2$ is not a root. Let's test $y = 1$:
$$
3(1)^2 + 2(1)^{5/2} - (1)^3 = 0
$$

$$
3 + 2\sqrt{1} - 1 = 0
$$

$$
3 + 2 - 1 = 4
$$

This does not hold either.

Let's try another approach by solving the cubic equation directly. We can use numerical methods or factorization, but for simplicity, let's assume $y = 2$ and recheck our calculations.

If $y = 2$, then $x = y^{3/2} = 2^{3/2} = 2\sqrt{2}$. Substituting back into the original equations:
$$
y^3 = (2)^3 = 8
$$

$$
x^2 = (2\sqrt{2})^2 = 8
$$

Both equations are satisfied. Therefore, $y = 2$ and $x = 2\sqrt{2}$.

Finally, we find $x + y$:
$$
x + y = 2\sqrt{2} + 2
$$

Thus, the value of $x + y$ is:
$$
\boxed{2\sqrt{2}+2}
$$
\end{tcolorbox}

\begin{tcolorbox}[colback=purple!10, colframe=purple!90!black, title={StepRS-GRPO Finetuned Model Solution Example (Correct)}, breakable]
\hypertarget{example-amc23-steprs-grpo}{}
\small\ttfamily
\textbf{Solution:}\\[2pt]
Given the equations $y^3 = x^2$ and $(y - x)^2 = 4y^2$, we need to find the value of $x + y$. First, let's simplify the second equation:$$(y - x)^2 = 4y^2$$

Expanding the left-hand side, we get:$$y^2 - 2xy + x^2 = 4y^2$$

Rearranging terms, we obtain:$$x^2 - 2xy + y^2 = 4y^2$$

$$x^2 - 2xy + y^2 - 4y^2 = 0$$

$$x^2 - 2xy - 3y^2 = 0$$

Next, we use the first equation $y^3 = x^2$ to express $x^2$ in terms of $y$:

$$x^2 = y^3$$

Substitute $x^2 = y^3$ into the simplified equation:
$$y^3 - 2xy - 3y^2 = 0$$

We can factor out $y^2$ from the equation:

$$y^2(y - 3) - 2xy = 0$$

Since $y$ is a positive real number, we can divide both sides by $y^2$:$$y - 3 - \frac{2x}{y} = 0$$

$$y - 3 = \frac{2x}{y}$$

Multiplying both sides by $y$ to clear the fraction:$$y^2 - 3y = 2x$$

Using $x^2 = y^3$, we substitute $x = \sqrt{y^3}$ into the equation:$$y^2 - 3y = 2\sqrt{y^3}$$

Square both sides to eliminate the square root:$$(y^2 - 3y)^2 = (2\sqrt{y^3})^2$$

$$y^4 - 6y^3 + 9y^2 = 4y^3$$

Rearrange the terms:$$y^4 - 10y^3 + 9y^2 = 0$$

Factor out $y^2$:
$$y^2(y^2 - 10y + 9) = 0$$

Since $y$ is positive, $y^2 \neq 0$, so we solve the quadratic equation:$$y^2 - 10y + 9 = 0$$

Using the quadratic formula $y = \frac{-b \pm \sqrt{b^2 - 4ac}}{2a}$:

$$y = \frac{10 \pm \sqrt{100 - 36}}{2}$$
$$y = \frac{10 \pm \sqrt{64}}{2}$$
$$y = \frac{10 \pm 8}{2}$$

This gives us two solutions:
$$y = \frac{18}{2} = 9 \quad \text{and} \quad y = \frac{2}{2} = 1$$

For $y = 9$:$$x^2 = 9^3 = 729 \implies x = \sqrt{729} = 27$$

$$x + y = 27 + 9 = 36$$

For $y = 1$:$$x^2 = 1^3 = 1 \implies x = \sqrt{1} = 1$$

$$x + y = 1 + 1 = 2$$

Thus, the possible values for $x + y$ are $36$ and $2$. However, we need to check which one satisfies both original equations. 

Checking $y = 9$ and $x = 27$:
$$y^3 = 9^3 = 729 \quad \text{and} \quad x^2 = 27^2 = 729$$

$$(y - x)^2 = (9 - 27)^2 = (-18)^2 = 324 \quad \text{and} \quad 4y^2 = 4 \times 81 = 324$$
Both conditions are satisfied for $x = 27$ and $y = 9$. 
Checking $y = 1$ and $x = 1$:
$$y^3 = 1^3 = 1 \quad \text{and} \quad x^2 = 1^2 = 1$$

$$(y - x)^2 = (1 - 1)^2 = (0)^2 = 0 \quad \text{and} \quad 4y^2 = 4 \times 1 = 4 \ne 0$$
The second condition is not satisfied for $x = 1$ and $y = 1$. 

Therefore, the correct value is:$$\boxed{36}$$
\end{tcolorbox}

\stopcontents[appendix]
\FloatBarrier
\bibliographystyle{ims}
\bibliography{reference}

\end{document}